\documentclass{article}

\usepackage[preprint]{neurips_2026}

\usepackage[utf8]{inputenc}
\usepackage[T1]{fontenc}
\usepackage[hyphens]{url}
\usepackage{graphicx}
\usepackage[table]{xcolor}
\def\UrlFont{\rm}
\usepackage{natbib}
\setcitestyle{round}
\usepackage{caption}
\usepackage{float}
\usepackage{placeins}
\usepackage{algorithm}
\usepackage{algorithmic}

\usepackage{newfloat}
\usepackage{listings}
\DeclareCaptionStyle{ruled}{labelfont=normalfont,labelsep=colon,strut=off}
\floatstyle{ruled}
\newfloat{listing}{tb}{lst}{}
\floatname{listing}{Listing}

\usepackage{booktabs}
\usepackage{longtable}
\usepackage{array}
\usepackage{calc}
\usepackage{multirow}
\usepackage{amsmath}
\usepackage{amsfonts}
\usepackage{nicefrac}
\usepackage{microtype}
\usepackage{hyperref}
\usepackage{simpleicons}
\usepackage{twemojis}

\definecolor{hfyellow}{HTML}{FFD21E}
\definecolor{linkblue}{HTML}{1A56B0}
\definecolor{citecolor}{HTML}{2980b9}
\definecolor{linkcolor}{HTML}{c0392b}
\hypersetup{
  colorlinks=true,
  citecolor=citecolor,
  linkcolor=linkcolor,
  urlcolor=linkblue,
}

\title{TLive-Omni: An Omni-Modal Understanding Model \\ for E-Commerce Live Streaming}

\author{
  TLive-Omni Team \\
  Taobao \& Tmall Group of Alibaba \\
}

\begin{document}

\setlength{\headheight}{21pt}

\makeatletter
\renewcommand{\@noticestring}{}
\makeatother

\maketitle

\begin{center}
\vspace{-30pt}
{\def\UrlFont{\ttfamily\color{linkcolor}}%
\begin{tabular}{@{}l@{}}
\raisebox{-0.2em}{\simpleicon{github}}~\url{https://github.com/TaoLiveAIGC/TLive-Omni} \\[3pt]
\raisebox{-0.2em}{\twemoji[height=1.1em]{1f917}}~\url{https://huggingface.co/TaoLiveAIGC/TLive-Omni-4B} \\[3pt]
\raisebox{-0.2em}{\twemoji[height=1.1em]{1f917}}~\url{https://huggingface.co/TaoLiveAIGC/TLive-Omni-9B}
\end{tabular}}
\end{center}

\begin{abstract}
% 介绍当前直播场景的痛点
E-commerce live streaming requires omni-modal understanding of noisy, temporally extended streams, where product facts are distributed across speech, video frames, product images, overlaid text, and user queries.
% 我们提出的模型
We present TLive-Omni, an omni-modal understanding model tailored to live-commerce scenarios. It maps image, video, audio, and text inputs into a unified representation space. For long-form live streaming analysis, we introduce Per-vGrid, a timestamped token organization that groups each video grid with its temporally corresponding audio within explicit boundary tokens to facilitate temporal alignment.
% 训练三阶段以及Faithful-RFT强化理解场景的训练方法
We design a three-stage supervised training recipe that progressively develops live-commerce understanding, from omni-modal perception to instruction-following responses. We then propose Faithful-RFT, a reinforcement fine-tuning stage that further improves answer faithfulness and expression quality while meeting real-time demands, scoring final responses directly with task-verifiable feedback rather than optimizing for reasoning-style exploration during rollout.
% 数据飞轮
Moreover, TLive-Omni is supported by a scenario-oriented atomic capability taxonomy and a compact data production engine that converts live-commerce audio, image, and video streams into training signals for speech recognition, speaker analysis, product visual grounding, text recognition, temporal grounding, video dense caption, and omni-modal QA, etc.
% 训练优化
For scalable training, a synchronized length-grouped sampler reduces padding while preserving comparable workloads across workers, while a lightweight dynamic sampling strategy regenerates rollout groups with near-zero reward variance to maintain meaningful relative advantages for GRPO.
% 实验内容
Experiments on e-commerce live streaming benchmarks demonstrate strong performance across live-commerce domain tasks, together with excellent generalization on general benchmarks.

\end{abstract}

\section{Introduction}
% 电商场景下的多模态理解难题
E-commerce live streaming poses a focused but challenging setting for omni-modal understanding. Product facts are distributed across host speech, video frames, product images, overlaid text, and user queries, while the supporting evidence may appear at different moments of a long stream. As a result, models must jointly interpret heterogeneous signals rather than process each modality in isolation, align audio and visual evidence to product-centric temporal segments, and express perception-derived answers faithfully for tasks such as automatic speech recognition, optical character recognition, product visual grounding, temporal grounding, and omni-modal question answering.
% 现有方法解决了什么，以及还有什么样的问题
General omni models and e-commerce-oriented systems have made important progress in this direction, but live-commerce understanding remains under-specified. Open-source omni models such as MiniCPM-o 4.5~\citep{minicpmo45}, Qwen3-Omni~\citep{qwen3omni}, OmniVinci~\citep{omnivinci}, and Nemotron 3 Nano Omni~\citep{nemotron3nano} demonstrate the feasibility of unified omni interaction, yet their data and evaluation are not primarily organized around product-centric live streaming understanding. Valley3 \citep{valley3} extends toward e-commerce scenarios. However, Valley3 is not primarily organized around fine-grained atomic capabilities for live streaming.

% 我们提出的方案，主要通过三阶段训练 + Faithful-RFT来全面提升点场直播场景理解
We present \textbf{TLive-Omni}, an omni-modal understanding model tailored to e-commerce live streaming. To couple heterogeneous modalities, TLive-Omni builds on a Qwen3.5 \citep{qwen35llm} backbone and integrates the pretrained audio encoder from Qwen3-Omni \citep{qwen3omni} into a unified interface. TLive-Omni supports up to 256K tokens of multimodal context, providing long-context capacity for extended live streaming segments. For audio--video alignment, we introduce Per-vGrid, which groups each video grid with the audio covering the same time interval in a span marked by explicit boundary tokens, together with a textual timestamp computed from the actual sampled frame indices. This explicit grid-level grouping keeps matched visual and audio evidence adjacent and makes their correspondence directly identifiable in the input sequence. Training begins with a three-stage supervised fine-tuning recipe that progressively develops live streaming understanding using both live-commerce and general multimodal supervision. 
% 由于直播视频理解希望有实时性要求，我们希望能快速给出答案/Response，在模型内隐式内化think traces，因此不显示优化think traces
Since perception-centered live-commerce tasks require answers that are both faithful to perceived evidence and timely enough for real-time live streaming, we introduce Faithful-RFT, a reinforcement fine-tuning stage that suppresses explicit think traces and scores final answers directly with task-verifiable rewards, improving answer faithfulness and expression quality for live-commerce understanding tasks. During rollout, a lightweight dynamic strategy resamples response groups with near-zero reward variance to yield higher-variance group-relative feedback, with vLLM~\citep{vllm} providing generation for this dynamic process. To support heterogeneous multimodal training across stages, our synchronized length-grouped sampling organizes mixed-modality batches with more compatible sequence lengths.

% 数据engine
TLive-Omni is organized around a scenario-oriented capability taxonomy and a compact data production engine. The taxonomy cover atomic capabilities across audio, image, video, and omni-modal understanding, including speech recognition, speaker analysis, product visual grounding, text recognition, temporal grounding, video dense caption, and omni-modal QA, etc. Based on this taxonomy, the data engine maps live-commerce streams into capability-specific supervision, enabling staged training data to support systematic improvement in live-commerce understanding. 

% 评估性能（业务数据以及通用数据） 我们的模型在业务数据表现很好，同时也在通用数据上表现不错
We further construct an in-house live-commerce evaluation suite to verify live streaming understanding capabilities across key dimensions. Experiments on in-house live-commerce, general-purpose multimodal and omni benchmarks indicate that TLive-Omni achieves strong performance on the business-oriented tasks while obtaining leading results on several general-purpose benchmarks, and remains competitive on the rest.

% \section{Related Work}
% \input{02_related_work}

\section{Architecture}
\subsection{Overview}
\label{sec:overview}

\begin{figure}[t]
\centering
\includegraphics[width=\textwidth]{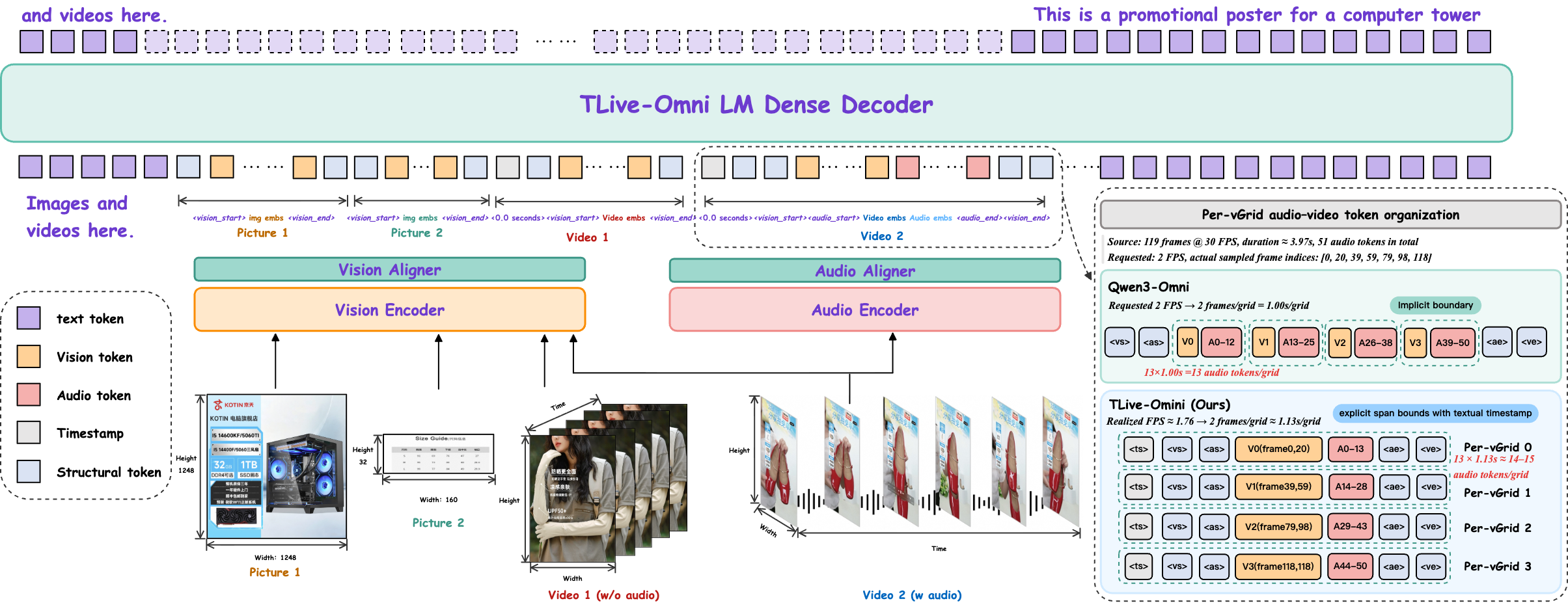}
\caption{The architectural overview of TLive-Omni. It is built upon a Qwen3.5 backbone and extended with the AuT audio encoder through a lightweight audio aligner. The inset illustrates the Per-vGrid token organization that groups each video grid with its temporally corresponding audio within explicit boundary tokens.}
\label{fig:architecture}
\end{figure}

% 模型形态定位：直播查询以文本指令为主，作答需要联合语音、画面、商品图与叠加文字的证据，因此模型形态为文本输出的全模态理解模型
TLive-Omni is a text-only output omni-modal understanding model for image, video, audio, and text inputs. Figure~\ref{fig:architecture} summarizes the architecture.
% 基座选型与嫁接决策：Qwen3.5 提供文本理解与指令遵循的中枢能力，视觉通路由基座原生 vision encoder 继承、保持不变；音频能力通过嫁接 Qwen3-Omni 的 AuT 音频理解塔补足
It uses a Qwen3.5 backbone~\citep{qwen35llm} as the language and vision substrate, and grafts the audio transformer (AuT) encoder from Qwen3-Omni~\citep{qwen3omni} into the same embedding space through a lightweight audio aligner. The staged training recipe then aligns the extended audio pathway with the backbone for live streaming understanding.

\subsection{Vision-Language Backbone}
\label{sec:arch-vision}

% 视觉-语言基座：Qwen3.5 提供文本理解、指令遵循和原生视觉处理；TLive-Omni 保留 text backbone
TLive-Omni uses Qwen3.5~\citep{qwen35llm} as its vision-language backbone, which provides the language-model substrate and native visual processing pipeline. After spatial merging, each image contributes $(h/32)\times(w/32)$ visual tokens, and each sampled video contributes $\lceil f/2\rceil\times(h/32)\times(w/32)$ visual tokens, where $f$ is the number of sampled frames and $h,w$ are the resized height and width, respectively. The native Qwen3.5 vision aligner applies a multi-layer perceptron (MLP) to map the merged visual features to the backbone embedding dimension.

\subsection{Audio Encoder}
\label{sec:arch-audio}

% 音频作为一等输入模态的动机：口播承载大量画面中不可见的商品事实；外部 ASR 级联会丢失音视频时间对应与说话人等副语言线索
In live commerce, host speech carries many product facts that are not visible in video frames. Transcribing speech with an external ASR system and feeding only the resulting text to the model would discard the temporal correspondence between speech and video, as well as paralinguistic cues such as speaker identity. TLive-Omni therefore keeps audio as a first-class input modality.
% AuT 音频塔规格；20M 小时预训练（Qwen3-Omni 报告口径)；13 tokens/s 压缩率让长音频在上下文预算内可负担
The audio encoder is the AuT adopted from Qwen3-Omni~\citep{qwen3omni}, trained from scratch on 20 million hours of audio data. It consumes 128-dimensional mel-spectrogram features at 16\,kHz, supports variable-length audio, and compresses speech into approximately 13 tokens per second, keeping long recordings computationally feasible within the context budget. A two-layer aligner projects the audio features into the backbone embedding space.

\subsection{Multimodal Temporal Alignment}
\label{sec:arch-alignment}

% 核心痛点：基座视频布局已提供可读时间戳，但直播理解还需要把同一时间片内的口播与画面证据放近，并在位置编码上绑定二者；朴素排布把 audio token 全部置于 video token 之后，口播与其指涉的画面证据在序列上完全错位
Long live streaming understanding requires explicit audio--video correspondence in the input token sequence. We introduce Per-vGrid, which organizes a video and its corresponding audio into a sequence of timestamped video grids. The visual content of a temporal grid and the audio segment covering the same time interval are placed in the same local span. Compared with Qwen3-Omni~\citep{qwen3omni}, Per-vGrid additionally prepends an explicit textual timestamp to each grid, makes the grid boundaries explicit, keeps each grid's video and audio tokens contiguous, and separates neighboring grids at the sequence level. This distinction is illustrated in Figure~\ref{fig:architecture}.

Per-vGrid further derives each grid's timestamp and audio span from the realized video sampling process. This distinction becomes visible when frame sampling involves integer rounding. As shown in Figure~\ref{fig:architecture}, consider a 119-frame source video at 30 FPS, whose duration is approximately 3.97 seconds. With a requested sampling rate of 2 FPS, the actual sampled frame indices can be $[0,20,39,59,79,98,118]$, giving a realized sampling rate of $7/119\times30\approx1.76$ FPS rather than exactly 2 FPS. Per-vGrid follows these actual sampled frames when assigning timestamps and audio spans. Since temporal patching groups two sampled frames into one grid and pads the final frame when necessary, the seven sampled frames form four grids, with each full grid spanning about $2/1.76\approx1.13$ seconds rather than the 1.0 seconds implied by the requested rate. With roughly 13 audio tokens per second, this changes the audio span of a full grid from about 13 tokens to about 14--15 tokens. Thus, when integer frame selection makes the realized sampling rate differ from the requested rate, both the grid timestamp and its audio-token span follow the actual sampled frames, preserving more precise temporal alignment.

\section{Supervised Fine-Tuning: Data and Recipe}
\subsection{Data Construction}
\label{sec:data-construction}

\begin{figure}[t]
\centering
\includegraphics[width=\textwidth]{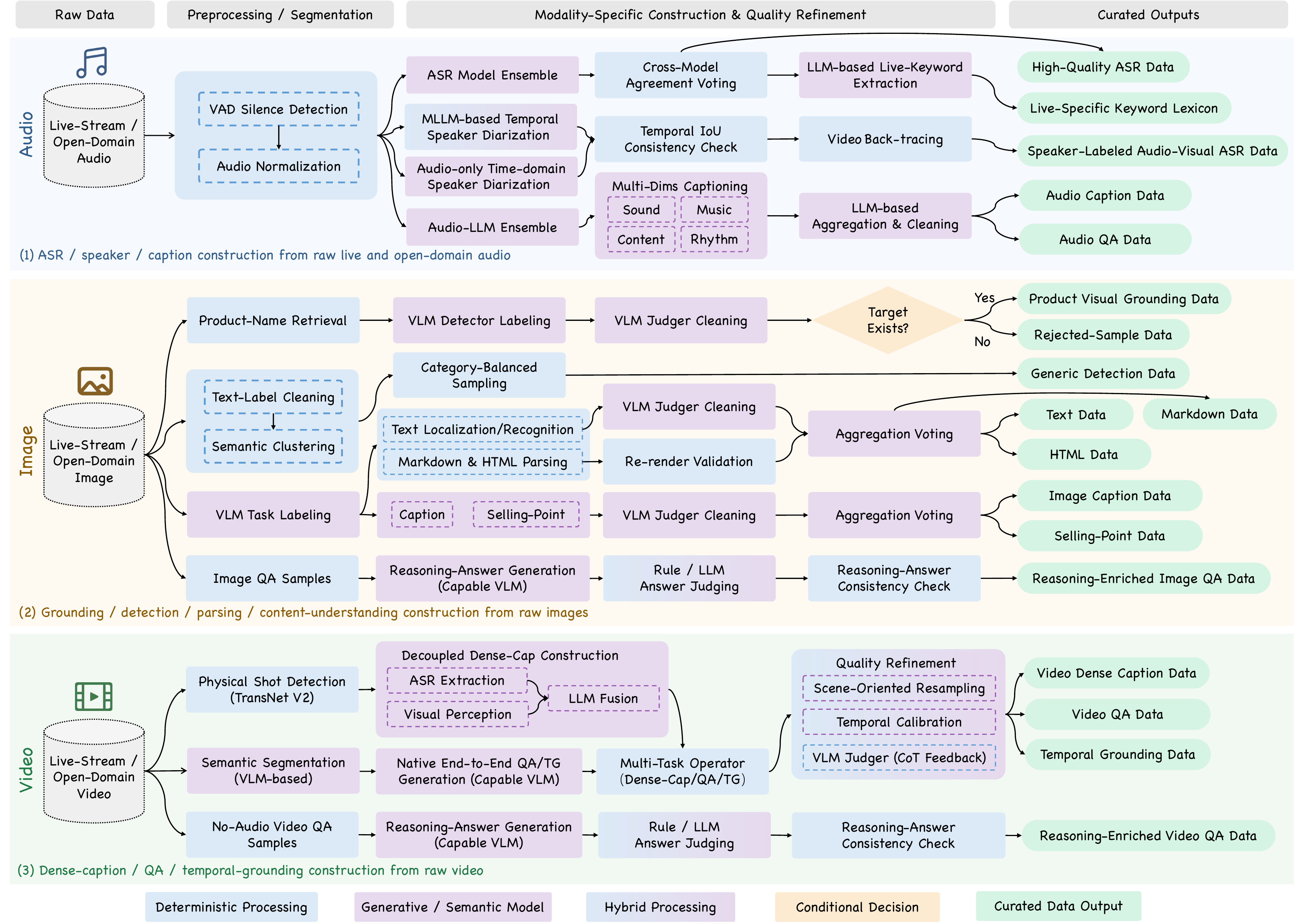}
\caption{TLive-Omni three-stage SFT data construction framework. Modality-specific generation and quality control transform audio, image, and video sources into task-grounded supervision for the three-stage SFT recipe.}
\label{fig:data-construction}
\end{figure}

Constructing supervision for omni-modal live-commerce understanding cannot rely on a single shared strategy: audio, image, and video sources exhibit distinct noise patterns and therefore require modality-specific construction. We process raw e-commerce data and curated general-domain data through separate audio, image, and video pathways, each with its own filtering and quality control, turning noisy inputs into task-grounded supervision. Figure~\ref{fig:data-construction} summarizes the source, construction, filtering, and output stages.

\paragraph{Audio pathway.}
Live-commerce speech poses two practical challenges for audio labeling. Hosts speak quickly and use domain-specific low-frequency terms such as brand names, materials, colors, and model numbers, which general ASR models may fail to recognize. Overlapping speakers and short interjections also make single-pass end-to-end diarization unreliable. 
After voice-activity detection (VAD) and audio normalization, we use cross-model agreement voting across an ASR model ensemble to obtain ASR pseudo-labels. A large language model (LLM) mines these transcripts for domain-specific low-frequency terms and builds a live-specific keyword lexicon for in-context ASR. 
For speaker-aware supervision, we cross-validate two independently generated speaker labels for the same segment: an audio-only time-domain diarization model assigns speaker identities from acoustic features alone, while a multimodal large language model (MLLM) predicts speaker identity from the ASR transcript together with visual cues. We compare the time spans from the two streams with a temporal intersection-over-union (IoU) consistency check, retaining high-overlap segments directly. In terms of mismatched segments, we use the corresponding video frames and lip-motion cues to verify the speaker assignment. 
Broader audio understanding is handled by separating captioning into sound, music, speech content, and speaker rhythm. Each dimension is generated by an audio-LLM ensemble and merged by an LLM into audio caption and audio QA-pair data.

\paragraph{Image pathway.}
The image pathway addresses challenges that are common in e-commerce imagery. 
Manual bounding-box annotation is expensive because product categories are visually diverse and live-stream backgrounds are cluttered. We construct product visual grounding data with a vision-language model (VLM) Detector--Judger loop. The detector proposes candidate boxes, the judger filters out inaccurate ones as rejected-sample data. 
For open-source generic-detection data, we cluster raw text labels into standardized categories and apply category-balanced sampling to reduce label noise and long-tail bias. 
In terms of Markdown and HTML parsing, we re-render the parsed content and compare it with the source image rather than trusting the VLM's output directly. 
Caption and selling-point data are filtered in the same source-consistency manner, with a VLM judger removing unsupported descriptions before aggregation. 
For reasoning-enriched Image QA, a capable VLM generates an answer with a reasoning trace for each selected Image QA sample. We then apply rule- or LLM-based judging to score the answer and keep only samples that pass judging and whose answer remains consistent with the trace.

\paragraph{Video pathway.}
Video annotation is complicated by the mismatch between physical shot boundaries and semantic event boundaries.
For dense captioning, TransNet V2~\citep{transnetv2} splits each video at physical shot boundaries, generating visually coherent clips. For each clip, a dedicated ASR model extracts the speech, while a VLM describes the visual content. An LLM combines the transcript and visual description into a dense caption.
For Video QA and temporal grounding, a VLM segments each video at semantic event boundaries, thereby keeping each clip focused on a self-contained semantic event. A capable VLM then directly generates QA pairs and temporal-grounding annotations from these clips.
The dense captions, QA pairs, and temporal-grounding annotations share a common refinement pipeline. Scene-oriented resampling improves coverage of long-tail scenarios, while temporal calibration adjusts clip boundaries to create targets of different durations. A VLM judger then checks factual and logical consistency as well as timestamp alignment, correcting or removing low-quality annotations.
For reasoning-enriched Video QA, we use QA samples from no-audio videos. A capable VLM generates an answer with a reasoning trace for each sample. We validate multiple-choice and numerical answers through rule-based matching and free-form answers with an LLM judge, retaining only samples that pass these checks.

\subsection{Three-Stage SFT Recipe}
\label{sec:training-recipe}

TLive-Omni follows a three-stage supervised fine-tuning (SFT) recipe that first establishes audio--language alignment, then strengthens audio understanding, and finally performs joint adaptation with audio, image, video, and text supervision. This progression separates modality alignment from capability learning and full multimodal adaptation, allowing each stage to update only the components required by its objective. \textbf{Stage~1} freezes the language model and audio encoder and trains only the audio aligner on 5M ASR samples, establishing an initial mapping from acoustic representations to the language-model embedding space.
\textbf{Stage~2} introduces a broader audio mixture comprising ASR, audio captioning, and audio QA over 26M audio samples. Training the audio encoder together with its aligner, while keeping the language model frozen, extends the audio pathway beyond transcription to speech content, sound events, music, and speaker-related cues.
\textbf{Stage~3} performs joint multimodal supervised fine-tuning over 14M multimodal samples spanning audio, image, video, and text data. The audio and visual encoders remain frozen, whereas their aligners and the language model are optimized for tasks including speech recognition, speaker analysis, product visual grounding,
text recognition, temporal grounding, video dense caption, omni-modal QA, and reasoning-enriched image and video QA. Detailed settings are reported in Appendix~\ref{app:sft-hparams}.

\subsection{Synchronized Length-Grouped Sampling}
\label{sec:length-group-sampling}

Training on heterogeneous multimodal data poses a practical batching challenge. Audio clips, product images, OCR-heavy pages, short videos, and long video segments vary substantially in token length and computational cost. Random sampling can place examples of disparate lengths in the same global batch, increasing padding and creating workload imbalance across workers.
Sequence packing reduces padding by concatenating short samples, but introduces a trade-off between fixed source-sample counts and full token-budget utilization. Suppose the memory budget is \(NL\) tokens, enough for \(N\) sequences of length \(L\). If each step is restricted to \(N\) source samples, packing several short samples into one sequence may produce only \(N'<N\) packed sequences, leaving part of the available token budget unused. Filling this budget with additional samples instead makes the number of source samples variable across steps, complicating control over the effective sample-level batch size and per-step data mixture. Packing also requires careful handling of block-diagonal attention, position IDs, loss masks, and multimodal metadata to preserve sample boundaries.

To address these issues, we propose a synchronized length-grouped sampler that forms fixed-size global batches without merging source samples. During initialization, it partitions samples by modality, sorts each partition by token length, and splits the sorted samples into global batches. This organization reduces padding while preserving a fixed sample count.
At each epoch, all workers use the same epoch-dependent seed. They therefore select the same modality and the same global batch at every step. Each selected global batch is partitioned into disjoint local batches, one per worker. Because samples within a global batch have similar lengths, these local batches impose comparable workloads across workers. During training, we enable the sampler with sequence packing disabled. Algorithm~\ref{alg:length-group-sampling} gives the detailed procedure.

\begin{algorithm}[t]
\caption{Synchronized length-grouped sampling}
\label{alg:length-group-sampling}
\renewcommand{\algorithmiccomment}[1]{\hfill\textit{// #1}}
\begin{algorithmic}[1]
\REQUIRE Dataset $\mathcal{D}$, modalities $\mathcal{M}$, world size $R$, local batch size $b$, seed $s$, epoch $e$, rank $r$
\STATE \textbf{Phase I: Registry construction (once per run)}
\STATE Set global batch size $B\leftarrow Rb$
\FOR{each modality $m\in\mathcal{M}$}
    \STATE Collect indices $\mathcal{I}_m$ and sort by token length \COMMENT{Reduce padding}
    \STATE Set $K_m\leftarrow\lfloor |\mathcal{I}_m|/B\rfloor$
    \STATE Form $\mathcal{Q}_m\leftarrow\{\mathcal{I}_m[kB:(k+1)B]\}_{k=0}^{K_m-1}$
    \STATE Discard $\mathcal{I}_m[K_mB:|\mathcal{I}_m|]$ \COMMENT{Keep global batch size fixed}
\ENDFOR
\STATE Make the registry $\{\mathcal{Q}_m\}_{m\in\mathcal{M}}$ identical across workers
\STATE \textbf{Phase II: Synchronized scheduling (each epoch $e$)}
\STATE Initialize every worker with seed $s+e$ \COMMENT{Identical random state}
\STATE Every worker builds the same shuffled copy $\widetilde{\mathcal{Q}}_m$ of each $\mathcal{Q}_m$
\WHILE{at least one $\widetilde{\mathcal{Q}}_m$ is nonempty}
    \STATE Sample $m$ in proportion to $|\widetilde{\mathcal{Q}}_m|$ \COMMENT{Number of remaining batches}
    \STATE Every worker pops the same global batch $\mathcal{G}_t$ from $\widetilde{\mathcal{Q}}_m$
    \STATE Worker $r$ yields $\mathcal{G}_t[rb:(r+1)b]$
\ENDWHILE
\end{algorithmic}
\end{algorithm}

\section{Faithful-RFT}
\label{sec:post-training}

The three-stage supervised recipe equips TLive-Omni with multimodal perception and task-solving capabilities, but its likelihood objective does not directly incorporate task-specific feedback on generated responses. In live-commerce applications, these responses must faithfully reflect perceived evidence while remaining timely for real-time understanding of live streams.
To meet these requirements, we introduce Faithful-RFT after the three-stage SFT. Faithful-RFT uses Group Relative Policy Optimization (GRPO)~\citep{deepseekmath} with task-verifiable rewards that directly score final responses. This approach improves answer faithfulness and expression quality without explicitly rewarding reasoning length or visible reasoning traces, thereby avoiding unnecessary generation overhead for real-time live streaming.

\subsection{Faithful-RFT Framework}
\label{sec:faithful-rft}

\paragraph{Initialization and data organization.}
Faithful-RFT starts from the Stage-3 model without a separate cold-start SFT stage. To preserve modality-specific schemas within a single optimization stage, we organize the data into four streams. We use $t_i$ to denote the task associated with an example, so that reward assignment can be formulated as task-conditioned routing rather than as a modality-level rule. The image stream includes representative tasks such as image QA, visual grounding, text recognition, and detailed captioning. The video-with-audio stream covers detailed captioning and Omni QA, while the video-without-audio stream focuses on video QA, shot understanding and temporal grounding. The audio stream includes ASR and audio QA. All streams are mixed within one GRPO stage rather than optimized as separate sequential stages.

\paragraph{Grouped rollout and optimization.}
For each multimodal prompt $x_i$, the policy samples a group of $G$ candidate responses $\{y_{i,g}\}_{g=1}^{G}$. The visual encoder, visual aligner, audio encoder, and audio aligner remain frozen during policy optimization. Following the outcome-supervision formulation of GRPO, all tokens in a response share the group-relative advantage below, where $R_{i,g}$ is the aggregated scalar reward defined in Eq.~\ref{eq:aggregated-reward}:
\begin{equation}
\widehat{A}_{i,g}
=
\frac{R_{i,g}-\operatorname{mean}_{g'}(R_{i,g'})}
{\operatorname{std}_{g'}(R_{i,g'})+\epsilon_s}.
\label{eq:group-relative-advantage}
\end{equation}
Let $\pi_\theta$, $\pi_{\theta_{\mathrm{old}}}$, and $\pi_{\mathrm{ref}}$ denote the policy being optimized, the rollout policy, and the frozen reference policy, respectively. In Eq.~\ref{eq:grpo-clipped-loss}, we use local group notation with $x=x_i$, $y_g=y_{i,g}$, $\widehat{A}_g=\widehat{A}_{i,g}$, and prefix $y_{g,<t}$. The implementation minimizes the group-level loss
\begin{equation}
\begin{aligned}
\mathcal{L}_i(\theta)
={}&\frac{1}{G}\sum_{g=1}^{G}\frac{1}{|y_g|}\sum_{t=1}^{|y_g|}
\Bigg[
-\min\!\left\{
\frac{\pi_\theta(y_{g,t}\mid x,y_{g,<t})}
{\pi_{\theta_{\mathrm{old}}}(y_{g,t}\mid x,y_{g,<t})}
\widehat{A}_{g},\right.\\
&\left.\operatorname{clip}\!\left(
\frac{\pi_\theta(y_{g,t}\mid x,y_{g,<t})}
{\pi_{\theta_{\mathrm{old}}}(y_{g,t}\mid x,y_{g,<t})},
1-\epsilon_l,1+\epsilon_h\right)\widehat{A}_{g}
\right\}
+\beta d^{\mathrm{KL}}_{g,t}
\Bigg].
\end{aligned}
\label{eq:grpo-clipped-loss}
\end{equation}
Here $d^{\mathrm{KL}}_{g,t}$ is the token-level KL penalty against the frozen reference policy under the same local notation. Faithful-RFT suppresses unnecessary explicit think traces, rather than encouraging visible reasoning traces or an additional thinking process. The corresponding optimization hyperparameters are reported in Appendix~\ref{app:faithful-rft-details}.

\paragraph{Rollout strategy.}
Rollout responses are generated by a vLLM engine deployed within the training recipe. Faithful-RFT extends the synchronized length-grouped sampler from Section~\ref{sec:length-group-sampling} with task-aware bucketing and repeated sampling. The sampler assigns each example to a bucket defined by its modality and task identifier $t_i$, so different tasks within the same modality are grouped separately. Within each bucket, examples are ordered by sequence length as in the sampler. This length grouping reduces input-side padding during rollout. For GRPO training, the repeated sampler yields each prompt index $G$ times, and the model generates the $G$ candidate responses. The repeated sampler also keeps a starvation counter over active buckets and prioritizes a bucket once it has remained unselected beyond the configured threshold.

During generation, we further use a lightweight dynamic resampling strategy to keep GRPO updates informative. Since the group-relative advantage in Eq.~\ref{eq:group-relative-advantage} depends on reward differences among the $G$ responses to the same prompt, a group whose aggregated rewards are all identical provides no relative preference signal. After scoring the rollout responses, the trainer computes the reward variance within each group, retains groups with nonzero variance, and regenerates candidate groups in cases of near-zero variance with adjusted generation settings or rewritten input prompts. This mechanism increases the proportion of groups that can produce meaningful relative advantages.

\subsection{Task-Conditioned Reward Function}
\label{sec:task-conditioned-reward}

The GRPO objective above requires a scalar reward \(R_{i,g}\) for each candidate response. In Faithful-RFT, this scalar reward is computed through task-conditioned reward routing, because a single multimodal batch can contain tasks with different notions of correctness. We use task-conditioned reward routing to select the reward functions applicable to each example. Each reward function $f_j$ declares an applicable task set $\mathcal{T}_j$ and is evaluated only when $t_i\in\mathcal{T}_j$. Inapplicable rewards return an invalid sentinel and are removed before reward aggregation.

\paragraph{Reward taxonomy.}
The reward pool is organized by evaluation mechanism and applied conditionally across the four data streams above. Rule-based rewards handle tasks with deterministic targets or machine-checkable structures, including multiple-choice questions, visual grounding and OCR. Some of these rewards use an LLM only to extract a final answer from a response that may contain an explanation or short thinking trace, and the extracted answer is still scored by a deterministic rule. LLM-judge rewards evaluate open-ended responses other free-form multimodal understanding tasks, for which exact string matching is insufficient. Finally, the final reward uses a lightweight format constraint to suppress unnecessary explicit think tags, without assigning reward to reasoning length, reasoning content, or visible chain-of-thought quality.

\paragraph{Reward aggregation.}
To obtain the scalar reward required by GRPO for each candidate, we exclude inapplicable or invalid reward outputs and renormalize the configured weights over the remaining rewards. Let $r_{i,g,j}$ be the score assigned by reward function $j$ to candidate $y_{i,g}$, and let $w_j$ be its configured weight. A reward is valid only when it is applicable to the example and returns a finite numerical score rather than the invalid sentinel:
\begin{equation}
v_{i,g,j}=\mathbf{1}\!\left[t_i\in\mathcal{T}_j\ \land\ \operatorname{valid}(r_{i,g,j})\right].
\label{eq:reward-validity}
\end{equation}
The configured weights are normalized over the valid rewards for each candidate:
\begin{equation}
\widetilde{w}_{i,g,j}=
\left\{
\begin{array}{ll}
\frac{v_{i,g,j}w_j}{\sum_k v_{i,g,k}w_k}, & \sum_k v_{i,g,k}w_k>0,\\
0, & \mathrm{otherwise}.
\end{array}
\right.
\label{eq:reward-weight-normalization}
\end{equation}
The scalar reward used by GRPO is then
\begin{equation}
R_{i,g}=\sum_{j:v_{i,g,j}=1} \widetilde{w}_{i,g,j}r_{i,g,j}.
\label{eq:aggregated-reward}
\end{equation}
As a result, each example is scored only by reward functions applicable to its task.
Together, the three-stage supervised recipe and Faithful-RFT yield two TLive-Omni variants: TLive-Omni-4B and TLive-Omni-9B.

\section{Evaluation}
\label{sec:evaluation}

We evaluate TLive-Omni model along two axes: live-commerce tasks that reflect the target application, and general benchmarks that measure the generalization capabilities.

\subsection{Live-Commerce Evaluation}
\label{sec:live-commerce-evaluation}

We evaluate the core multimodal capabilities required for live-commerce understanding. The suite is built from live-commerce sources and covers speech transcription, speaker-attributed ASR, audio description and question answering, product visual grounding, text localization/recognition/classification, temporal grounding, dense video caption, video question answering, and shot understanding. Appendix~\ref{app:evaluation-metrics} defines the detailed evaluation metrics used in these tasks.

\begin{table}[!htbp]
\centering
\small
\setlength{\tabcolsep}{4pt}
\resizebox{\textwidth}{!}{%
\begin{tabular}{@{}lcccccc@{}}
\toprule
\multicolumn{1}{c}{\multirow{2}{*}{\textbf{Model}}} & \multirow{2}{*}{\textbf{Params}} & \textbf{Live-Commerce ASR} & \textbf{Speaker-Attributed ASR} & \multicolumn{2}{c}{\textbf{Audio Description}} & \textbf{Audio QA} \\
\cmidrule(lr){3-3}\cmidrule(lr){4-4}\cmidrule(lr){5-6}\cmidrule(lr){7-7}
 & & \textbf{CER $\downarrow$} & \textbf{cpWER $\downarrow$} & \textbf{Acc. $\uparrow$} & \textbf{Hal. $\downarrow$} & \textbf{Acc. $\uparrow$} \\
\midrule
\multicolumn{7}{@{}l}{\textbf{\textit{Closed-source Omni models}}} \\
Gemini 2.5 Flash & - & 16.30 & 17.14 & 65.21 & 26.19 & 76.28 \\
Gemini 2.5 Pro & - & 11.48 & 12.17 & 81.10 & 14.16 & 82.85 \\
Gemini 3 Flash & - & 15.18 & 19.04 & 68.27 & 26.17 & 74.68 \\
Gemini 3 Pro & - & 12.09 & 11.67 & 85.07 & 10.92 & 88.62 \\
Gemini 3.5 Flash & - & 13.09 & 11.99 & 79.97 & 14.36 & 87.99 \\
Qwen3.5-Omni Flash & - & 6.81 & 13.23 & 62.82 & 27.81 & 78.04 \\
\midrule
\multicolumn{7}{@{}l}{{\textbf{\textit{Open-source Audio models}}}} \\
MiMo-Audio & 7B & 12.71 & -- & 64.26 & 26.01 & 70.97 \\
Fun-Audio-Chat & 8B & 14.55 & -- & 61.35 & 32.21 & 69.71 \\
Step-Audio-R1.1 & 32B & 10.21 & -- & 75.08 & \textbf{20.50} & 69.80 \\
\midrule
\multicolumn{7}{@{}l}{{\textbf{\textit{Open-source Omni models}}}} \\
OmniVinci & 9B & -- & -- & 39.90 & 47.36 & 66.51 \\
Nemotron 3 Nano Omni & 30B-A3B & 12.10 & 17.65 & 33.01 & 39.77 & 64.90 \\
Ming-Lite-Omni v1.5 & 20B-A3B & 10.06 & -- & 45.99 & 44.05 & 40.54 \\
MiniCPM-o 2.6 & 8B & 13.88 & -- & 49.84 & 41.76 & 39.74 \\
MiniCPM-o 4.5 & 9B & 10.70 & 18.89 & 47.59 & 52.41 & 42.47 \\
Qwen2.5-Omni & 7B & 7.86 & -- & 47.92 & 36.92 & 61.38 \\
Qwen3-Omni & 30B-A3B & 6.75 & 27.84 & 61.06 & 30.22 & \textbf{76.76} \\
\midrule
\multicolumn{7}{@{}l}{\textbf{\textit{Ours}}} \\
TLive-Omni & 4B & \underline{6.66} & \underline{12.88} & \textbf{76.12} & \underline{20.97} & 72.60 \\
TLive-Omni & 9B & \textbf{6.46} & \textbf{12.27} & \underline{75.96} & 21.00 & \underline{76.28} \\
\bottomrule
\end{tabular}%
}
\caption{Live-commerce audio evaluation covering live-commerce ASR, speaker-attributed ASR, audio description and question answering. A dash denotes an unreported result or undisclosed parameter count. The Best results among open-source models are marked in \textbf{bold}, while the second-best results are in \underline{underlined}.}
\label{tab:live-audio-results}
\end{table}

We compare against open-source baselines from the MiniCPM-family models~\citep{minicpmo26,minicpmo45}, Ming-Lite-Omni~\citep{mingomni}, OmniVinci~\citep{omnivinci}, Nemotron 3 Nano Omni~\citep{nemotron3nano}, Qwen-Omni-family models~\citep{qwen25omni,qwen3omni,qwen35omni}, Step-Audio~\citep{stepaudior11}, Fun-Audio-Chat~\citep{funaudiochat}, and MiMo-Audio~\citep{mimoaudio}, together with Gemini-family models~\citep{gemini,gemini25,gemini3flash,gemini3pro,gemini35flash}.

As shown in Table~\ref{tab:live-audio-results}, we evaluate live-commerce ASR using character error rate (CER) and speaker-attributed ASR using concatenated minimum-permutation word error rate (cpWER), which minimizes the total error over speaker assignments. For audio description, Accuracy (Acc.) evaluates whether generated descriptions support correct answers to audio-grounded questions, while Hallucination Rate (Hal.) reports the rate of unsupported content in those descriptions. Audio-QA Accuracy measures audio question answering under rule-based answer matching. On ASR, TLive-Omni-9B achieves the lowest CER, with TLive-Omni-4B close behind. Their cpWER scores are among the lower reported results, showing strong performance on speaker-attributed ASR for live-stream.

Table~\ref{tab:live-image-results} shows product-centric image understanding including product visual grounding and text understanding. We evaluate product visual grounding with Average Precision at an IoU threshold of 0.5 (AP@IoU=0.5) in both live-stream frames (Live) and product images (Prod).
For text understanding, we report text localization F1 score (Loc. F1), normalized edit distance for text recognition (Rec. NED), and text classification accuracy over commerce-oriented semantic labels (Cls. Acc.). Rec. NED is reported as a percentage.
The two TLive-Omni variants achieve the highest Prod AP, text localization, and classification scores, as well as the lowest recognition edit distances among the evaluated open-source and closed-source models, while also remaining competitive with Gemini 3.5 Flash~\citep{gemini35flash}, the strongest closed-source model on Live AP.

\begin{table}[!htbp]
\centering
\small
\setlength{\tabcolsep}{4pt}
\resizebox{\textwidth}{!}{%
\begin{tabular}{@{}lccccccc@{}}
\toprule
\multicolumn{1}{c}{\multirow{2}{*}{\textbf{Model}}} & \multirow{2}{*}{\textbf{Params}} & \multicolumn{2}{c}{\textbf{Visual Grounding}} & \multicolumn{3}{c}{\textbf{Text Understanding}} \\
\cmidrule(lr){3-4}\cmidrule(lr){5-7}
 & & \textbf{Live AP $\uparrow$} & \textbf{Prod AP $\uparrow$} & \textbf{Loc. F1 $\uparrow$} & \textbf{Rec. NED $\downarrow$} & \textbf{Cls. Acc. $\uparrow$} \\
\midrule
\multicolumn{7}{@{}l}{\textbf{\textit{Closed-source Omni models}}} \\
Gemini 2.5 Flash & - & 61.08 & 28.81 & 20.52 & 43.28 & 51.21 \\
Gemini 2.5 Pro & - & 51.98 & 32.63 & 31.60 & 27.82 & 61.86 \\
Gemini 3 Flash & - & 80.38 & 65.67 & 61.11 & 16.25 & 69.11 \\
Gemini 3 Pro & - & 73.80 & 58.83 & 68.60 & 9.72 & 76.86 \\
Gemini 3.5 Flash & - & 84.15 & 74.89 & 64.44 & 16.64 & 69.76 \\
Qwen3.5-Omni Flash & - & 79.96 & 60.44 & 74.07 & 12.48 & 53.25 \\
\midrule
\multicolumn{7}{@{}l}{{\textbf{\textit{Open-source Omni models}}}} \\
OmniVinci & 9B & 34.86 & 8.93 & 50.25 & 32.77 & 57.29 \\
Nemotron 3 Nano Omni & 30B-A3B & 73.08 & 48.62 & 52.91 & 29.42 & 37.86 \\
Ming-Lite-Omni v1.5 & 20B-A3B & 52.46 & 40.73 & 13.27 & 59.16 & 32.94 \\
MiniCPM-o 2.6 & 8B & 3.82 & 1.77 & 5.74 & 77.58 & 15.92 \\
MiniCPM-o 4.5 & 9B & 23.90 & 53.63 & 5.43 & 71.65 & 11.62 \\
Qwen2.5-Omni & 7B & 75.61 & 22.85 & 42.64 & 37.79 & 51.25 \\
Qwen3-Omni & 30B-A3B & 79.22 & 68.88 & 30.46 & 14.83 & 69.46 \\
\midrule
\multicolumn{7}{@{}l}{\textbf{\textit{Ours}}} \\
TLive-Omni & 4B & \textbf{82.85} & \textbf{91.45} & \underline{86.99} & \underline{4.72} & \underline{79.06} \\
TLive-Omni & 9B & \underline{82.33} & \underline{89.96} & \textbf{87.59} & \textbf{4.24} & \textbf{79.85} \\
\bottomrule
\end{tabular}%
}
\caption{Live-commerce image evaluation covering product visual grounding and text understanding. A dash denotes an undisclosed parameter count. The Best results among open-source models are marked in \textbf{bold}, while the second-best results are in \underline{underlined}.}
\label{tab:live-image-results}
\end{table}

Table~\ref{tab:live-video-results} reports video understanding abilities across temporal grounding, dense video caption, video question answering, and shot understanding.
We use mIoU to evaluate the temporal grounding (TG) ability of product or event intervals in live-stream videos. When an example contains multiple intervals, scoring uses interval-level matching rather than a single union IoU over all segments. Dense Caption Accuracy and Hallucination Rate report the correctness and unsupported-content rate of time-aware dense video descriptions, while Video QA Accuracy reports question answering accuracy over video evidence. Shot understanding is evaluated across four structured perspectives: layout, shot size, camera angle, and content category. TLive-Omni-9B achieves the highest TG mIoU, Video QA Accuracy, Dense Caption Accuracy, and the lowest Hallucination Rate, while TLive-Omni-4B obtains the second-best open-source results on the same four metrics. For shot understanding, TLive-Omni-9B ranks first in camera angle and content category among open-source models, while TLive-Omni-4B ranks second in shot size and content category. These results highlight the advantage of TLive-Omni in live video understanding.

\begin{table}[!htbp]
\centering
\small
\setlength{\tabcolsep}{3pt}
\resizebox{\textwidth}{!}{%
\begin{tabular}{@{}lccccccccc@{}}
\toprule
\multicolumn{1}{c}{\multirow{2}{*}{\textbf{Model}}} & \multirow{2}{*}{\textbf{Params}} & \textbf{TG} & \multicolumn{2}{c}{\textbf{Dense Caption}} & \textbf{Video QA} & \multicolumn{4}{c}{\textbf{Shot Understanding}} \\
\cmidrule(lr){3-3}\cmidrule(lr){4-5}\cmidrule(lr){6-6}\cmidrule(lr){7-10}
 & & \textbf{mIoU $\uparrow$} & \textbf{Acc. $\uparrow$} & \textbf{Hal. $\downarrow$} & \textbf{Acc. $\uparrow$} & \textbf{Layout $\uparrow$} & \textbf{Shot Size $\uparrow$} & \textbf{Camera $\uparrow$} & \textbf{Content $\uparrow$} \\
\midrule
\multicolumn{10}{@{}l}{\textbf{\textit{Closed-source Omni models}}} \\
Gemini 2.5 Flash & - & 76.50 & 54.60 & 10.97 & 88.21 & 80.00 & 46.80 & 84.20 & 68.60 \\
Gemini 2.5 Pro & - & 76.22 & 41.95 & 16.88 & 92.62 & 85.20 & 50.80 & 76.00 & 70.40 \\
Gemini 3 Flash & - & 77.43 & 32.21 & 20.76 & 89.64 & 76.80 & 45.70 & 78.50 & 71.60 \\
Gemini 3 Pro & - & 77.90 & 37.80 & 20.99 & 84.36 & 80.40 & 43.40 & 75.70 & 74.80 \\
Gemini 3.5 Flash & - & 77.90 & 33.80 & 17.30 & 86.90 & 83.40 & 44.20 & 75.50 & 70.20 \\
Qwen3.5-Omni Flash & - & 62.10 & 32.94 & 20.91 & 87.28 & 84.40 & 48.90 & 85.50 & 66.40 \\
\midrule
\multicolumn{10}{@{}l}{{\textbf{\textit{Open-source Omni models}}}} \\
OmniVinci & 9B & 13.10 & 18.59 & 27.13 & 72.51 & 73.60 & \textbf{52.70} & 68.10 & 49.20 \\
Nemotron 3 Nano Omni & 30B-A3B & 23.39 & 17.96 & 16.62 & 82.56 & \underline{79.20} & 34.00 & 80.20 & 58.60 \\
Ming-Lite-Omni v1.5 & 20B-A3B & 14.34 & 13.81 & 39.33 & 64.51 & 74.60 & 41.70 & 72.80 & 51.60 \\
MiniCPM-o 2.6 & 8B & 14.56 & 10.53 & 26.93 & 60.30 & 66.60 & 38.30 & 68.30 & 49.00 \\
MiniCPM-o 4.5 & 9B & 43.20 & 21.06 & 28.61 & 84.62 & 78.20 & 42.80 & 79.20 & 66.60 \\
Qwen2.5-Omni & 7B & 30.83 & 16.51 & 36.44 & 75.48 & 74.40 & 38.10 & \underline{81.20} & 68.00 \\
Qwen3-Omni & 30B-A3B & 39.22 & 21.44 & 25.82 & 81.62 & \textbf{82.20} & 37.40 & 76.10 & 63.60 \\
\midrule
\multicolumn{10}{@{}l}{\textbf{\textit{Ours}}} \\
TLive-Omni & 4B & \underline{77.63} & \underline{69.23} & \underline{9.57} & \underline{92.31} & 78.40 & \underline{51.20} & 80.90 & \underline{69.80} \\
TLive-Omni & 9B & \textbf{81.49} & \textbf{74.63} & \textbf{8.76} & \textbf{93.23} & 77.00 & 51.00 & \textbf{82.00} & \textbf{71.00} \\
\bottomrule
\end{tabular}%
}
\caption{Live-commerce video evaluation covering temporal grounding, dense video caption, video question answering, and shot understanding. A dash denotes an unreported result or undisclosed parameter count. The Best results among open-source models are marked in \textbf{bold}, while the second-best results are in \underline{underlined}.}
\label{tab:live-video-results}
\end{table}

\subsection{General Benchmark Evaluation}
\label{sec:general-benchmark-evaluation}

General benchmarks evaluate whether a model retains broad multimodal capabilities beyond the target live-commerce domain. This evaluation is important because vertical-domain specialization can improve target-domain performance while weakening general reasoning, perception, or cross-modal understanding, thereby narrowing the model's usable scenarios. We therefore evaluate TLive-Omni across image reasoning and question answering, hallucination/OCR/grounding/spatial reasoning, video understanding and temporal grounding, and omni-modal perception and reasoning. The results show that TLive-Omni maintains strong generalization across these general multimodal benchmarks, improves over Qwen3.5~\citep{qwen35llm} 4B and 9B backbones on the majority of these benchmarks while retaining and eliciting broad omni-modal understanding capabilities.

All models are evaluated in their instruction-tuned or chat variants, without dedicated reasoning modes. The general benchmark comparisons include Qwen-family models~\citep{qwen3vl,qwen35llm,qwen25omni,qwen3omni,qwen35omni}, MiniCPM-family models~\citep{minicpmo26,minicpmo45,minicpmv45}, InternVL3.5~\citep{internvl35}, NVILA~\citep{nvila}, MiMo-VL~\citep{mimovl}, Ming-Lite-Omni~\citep{mingomni}, InteractiveOmni~\citep{interactiveomni}, VITA-1.5~\citep{vita15}, Valley-family models~\citep{valley3,valley25}, SAIL-VL2~\citep{sailvl2}, LLaVA-OneVision-family models~\citep{llavaonevision,llavaonevision2}, LLaVA-Video~\citep{llavavideo}, LongVU~\citep{longvu}, LongVILA~\citep{longvila}, Kangaroo~\citep{kangaroo}, Video-XL-2~\citep{videoxl2}, VideoLLaMA 3~\citep{videollama3}, VideoChat3~\citep{videochat3}, Molmo2~\citep{molmo2}, Mage-VL~\citep{magevl}, OmniVinci~\citep{omnivinci}, Nemotron 3 Nano Omni~\citep{nemotron3nano}, video-SALMONN 2~\citep{videosalmonn2}, GPT-4o~\citep{gpt4o}, GPT-5~\citep{gpt5syscard}, and Gemini-family models~\citep{gemini,gemini25,gemini3pro,gemini31pro}. The evaluation prompts are reported in Appendix~\ref{app:general-benchmark-prompts}.

Table~\ref{tab:general-image-results} focuses on image-centric reasoning and question answering. MMMU~\citep{mmmu} assesses multimodal understanding and reasoning with domain-specific knowledge, while MathVista~\citep{mathvista} evaluates mathematical reasoning in visual contexts. DynaMath~\citep{dynamath} measures mathematical reasoning robustness under visual and textual variations of the same problem, whereas VLMsAreBlind~\citep{vlmsareblind} evaluates low-level visual perception requiring precise spatial information. MMBench~\citep{mmbench} and MMStar~\citep{mmstar} provide broad coverage of multimodal perception and reasoning. RealWorldQA~\citep{realworldqa} evaluates spatial and physical understanding of everyday scenes, while SimpleVQA~\citep{simplevqa} measures factuality in short-answer visual question answering. Compared with open-source baselines, TLive-Omni achieves the best results on MMBench and RealWorldQA, and ranks second on MMMU, MathVista, DynaMath, VLMsAreBlind, MMStar, and SimpleVQA.

\begin{table}[!htbp]
\centering
\tiny
\setlength{\tabcolsep}{2.2pt}
\resizebox{\textwidth}{!}{%
\begin{tabular}{@{}lccccccccc@{}}
\toprule
\multicolumn{1}{c}{\textbf{Model}} & \textbf{Params} & \textbf{MMMU} & \textbf{MathVista} & \textbf{DynaMath} & \textbf{VLMsAreBlind} & \textbf{MMBench} & \textbf{RealWorldQA} & \textbf{MMStar} & \textbf{SimpleVQA} \\
\midrule
\multicolumn{10}{@{}l}{\textbf{\textit{Closed-source models}}} \\
Gemini 2.5 Flash & - & 76.3 & 75.3 & 69.7 & 75.9 & 86.6 & 75.7 & 75.8 & 59.2 \\
Gemini 2.5 Pro & - & 80.9 & 77.7 & 78.5 & 78.5 & 88.4 & 76.0 & 78.5 & 66.9 \\
Gemini 3 Pro & - & 87.2 & 87.9 & 85.1 & -- & 93.7 & 83.3 & 83.1 & 73.2 \\
GPT-4o & - & 70.7 & 63.8 & 54.4 & -- & 86.0 & -- & -- & -- \\
GPT-5 (minimal) & - & 74.4 & 50.9 & 74.0 & 53.4 & 81.3 & 77.3 & 65.2 & 56.7 \\
Qwen3.5-Omni Flash & - & 76.9 & 82.9 & 79.3 & -- & 88.8 & 77.5 & 75.7 & 54.4 \\
\midrule
\multicolumn{10}{@{}l}{\textbf{\textit{Open-source VLM models}}} \\
MiMo-VL-SFT & 7B & 64.6 & 81.8 & 46.9 & \textbf{78.0} & 84.5 & -- & -- & -- \\
SAIL-VL2 & 8B & 55.4 & 76.4 & 17.8 & -- & -- & 76.3 & 70.7 & -- \\
Valley2.5 & 8B & 62.1 & 74.4 & 32.7 & -- & 85.5 & 70.5 & 67.3 & -- \\
LLaVA-OneVision-2 & 8B & -- & -- & -- & -- & 85.7 & 69.7 & 64.8 & -- \\
InternVL3.5 & 4B & 66.6 & 77.1 & 35.7 & -- & 80.3 & 66.3 & 65.0 & -- \\
InternVL3.5 & 8B & \underline{73.4} & 78.4 & 37.7 & -- & 79.5 & 67.5 & 69.3 & -- \\
Qwen3-VL & 4B & 67.4 & 73.7 & 65.3 & 71.9 & 83.9 & 70.9 & 69.8 & 48.0 \\
Qwen3-VL & 8B & 69.6 & 77.2 & 67.7 & 74.0 & 84.5 & 71.5 & 70.9 & \textbf{50.2} \\
Qwen3.5 & 4B & 72.1 & 81.0 & 69.6 & 62.3 & 86.3 & 72.5 & 74.8 & 44.6 \\
Qwen3.5 & 9B & \textbf{74.2} & \textbf{82.2} & \textbf{74.6} & 71.8 & \underline{87.7} & 72.9 & \textbf{76.3} & 48.9 \\
\midrule
\multicolumn{10}{@{}l}{\textbf{\textit{Open-source Omni models}}} \\
InteractiveOmni & 4B & 61.1 & 61.7 & -- & -- & 78.9 & -- & 62.6 & -- \\
InteractiveOmni & 8B & 66.9 & 68.0 & -- & -- & 81.4 & -- & 66.8 & -- \\
VITA-1.5 & 7B & 52.1 & 66.2 & -- & -- & 76.7 & -- & 59.9 & -- \\
Valley3 & 8B & 69.3 & -- & -- & -- & -- & -- & -- & -- \\
OmniVinci & 9B & 49.7 & 63.5 & -- & -- & -- & 67.5 & -- & -- \\
Nemotron 3 Nano Omni & 30B-A3B & 55.2 & 71.9 & -- & -- & -- & -- & -- & -- \\
Ming-Lite-Omni v1.5 & 20B-A3B & 54.3 & 72.0 & -- & -- & -- & -- & 65.1 & -- \\
MiniCPM-o 2.6 & 8B & 50.4 & 71.9 & -- & -- & 80.5 & -- & 64.0 & -- \\
MiniCPM-o 4.5 & 9B & 67.6 & -- & -- & -- & 87.6 & -- & 73.1 & -- \\
Qwen2.5-Omni & 7B & 59.2 & 67.9 & -- & -- & 81.8 & 70.3 & 64.0 & -- \\
Qwen3-Omni & 30B-A3B & 69.1 & 75.9 & -- & -- & -- & -- & 68.5 & -- \\
\midrule
\multicolumn{10}{@{}l}{\textbf{\textit{Ours}}} \\
TLive-Omni & 4B & 70.9 & 79.9 & 72.5 & 71.8 & 87.0 & \textbf{77.7} & 73.9 & 47.6 \\
TLive-Omni & 9B & \underline{73.4} & \underline{81.9} & \underline{73.3} & \underline{75.5} & \textbf{88.9} & \underline{76.6} & \underline{75.1} & \underline{50.0} \\
\bottomrule
\end{tabular}%
}
\caption{General image benchmark results on MMMU, MathVista, DynaMath, VLMsAreBlind, MMBench, RealWorldQA, MMStar, and SimpleVQA. MMBench results are reported on the EN-DEV-v1.1 split. A dash denotes an unreported result or undisclosed parameter count. The Best results among open-source models are marked in \textbf{bold}, while the second-best results are in \underline{underlined}.}
\label{tab:general-image-results}
\end{table}

Table~\ref{tab:general-image-ocr-results} extends image evaluation to hallucination, diagram and chart understanding, OCR perception, visual grounding, embodied reasoning, and spatial reasoning. It includes HallusionBench~\citep{hallusionbench} for hallucination, AI2D~\citep{ai2d} and CharXiv~\citep{charxiv} for diagram and chart understanding, OCRBench~\citep{ocrbench} and CC-OCR~\citep{ccocr} for OCR perception, and RefCOCO~\citep{refcoco}, ERQA~\citep{geminirobotics}, and EmbSpatialBench~\citep{embspatialbench} for grounding, embodied reasoning, and spatial reasoning. These results indicate that TLive-Omni preserves competitive general image reasoning while showing particular strength on hallucination, OCR-centric perception, and spatial reasoning.

\begin{table}[!htbp]
\centering
\tiny
\setlength{\tabcolsep}{2.2pt}
\resizebox{\textwidth}{!}{%
\begin{tabular}{@{}lccccccccc@{}}
\toprule
\multicolumn{1}{c}{\textbf{Model}} & \textbf{Params} & \textbf{Hallusion} & \textbf{AI2D} & \textbf{OCRBench} & \textbf{CC-OCR} & \textbf{CharXiv(RQ)} & \textbf{RefCOCO} & \textbf{ERQA} & \textbf{EmbSpatial} \\
\midrule
\multicolumn{10}{@{}l}{\textbf{\textit{Closed-source models}}} \\
Gemini 2.5 Flash & - & 59.1 & 87.7 & 86.4 & 74.8 & 60.1 & -- & -- & -- \\
Gemini 2.5 Pro & - & 60.9 & 90.0 & 87.2 & 76.8 & 62.9 & -- & 50.3 & 73.3 \\
Gemini 3 Pro & - & 68.6 & 94.1 & 90.4 & 79.0 & 81.4 & 84.1 & 70.5 & 61.2 \\
GPT-4o & - & -- & 82.6 & 84.3 & -- & -- & -- & -- & -- \\
GPT-5 (minimal) & - & 53.7 & 84.1 & 78.7 & 66.1 & 57.8 & -- & 42.0 & 75.1 \\
Qwen3.5-Omni Flash & - & -- & 89.0 & 89.1 & 80.8 & 64.4 & 92.6 & 50.0 & 82.7 \\
\midrule
\multicolumn{10}{@{}l}{\textbf{\textit{Open-source VLM models}}} \\
MiMo-VL-SFT & 7B & -- & 83.2 & 87.6 & -- & 54.4 & 85.7 & -- & -- \\
SAIL-VL2 & 8B & 55.1 & 87.7 & \textbf{91.3} & -- & -- & 74.0 & -- & -- \\
Valley2.5 & 8B & 56.3 & 84.4 & 87.0 & -- & -- & -- & -- & -- \\
LLaVA-OneVision-2 & 8B & -- & 84.3 & 78.2 & -- & -- & -- & 43.3 & 78.1 \\
InternVL3.5 & 4B & 44.8 & 82.6 & 82.2 & -- & 39.6 & 89.4 & 38.5 & -- \\
InternVL3.5 & 8B & 54.5 & 84.0 & 84.0 & -- & 44.4 & \underline{89.7} & 41.0 & 73.2 \\
Qwen3-VL & 4B & 57.6 & 84.1 & 88.1 & 76.2 & 39.7 & 89.0 & 41.3 & \underline{79.6} \\
Qwen3-VL & 8B & 61.1 & 85.7 & 89.6 & 79.9 & 46.4 & 89.1 & 45.8 & 78.5 \\
Qwen3.5 & 4B & \underline{76.9} & 87.1 & 85.9 & 71.1 & 62.9 & 87.6 & 46.8 & 76.6 \\
Qwen3.5 & 9B & 76.0 & 88.0 & 88.5 & 73.4 & \textbf{67.5} & \textbf{90.0} & \underline{47.3} & 78.7 \\
\midrule
\multicolumn{10}{@{}l}{\textbf{\textit{Open-source Omni models}}} \\
InteractiveOmni & 4B & 52.2 & 83.8 & 80.0 & -- & -- & -- & -- & -- \\
InteractiveOmni & 8B & 61.3 & 84.3 & 83.7 & -- & -- & -- & -- & -- \\
VITA-1.5 & 7B & 44.9 & 79.3 & 73.2 & -- & -- & -- & -- & -- \\
Valley3 & 8B & 55.9 & -- & -- & -- & -- & -- & -- & -- \\
Nemotron 3 Nano Omni & 30B-A3B & -- & \underline{88.5} & 88.3 & -- & 49.1 & 80.6 & -- & -- \\
Ming-Lite-Omni v1.5 & 20B-A3B & 54.6 & 84.9 & 88.9 & -- & -- & 87.8 & -- & -- \\
MiniCPM-o 2.6 & 8B & 51.9 & 85.8 & 89.7 & -- & -- & -- & -- & -- \\
MiniCPM-o 4.5 & 9B & 63.2 & 87.6 & 87.6 & -- & -- & -- & -- & -- \\
Qwen2.5-Omni & 7B & -- & 83.2 & -- & -- & -- & 87.7 & -- & -- \\
Qwen3-Omni & 30B-A3B & 59.7 & 85.2 & 86.0 & -- & 61.1 & -- & -- & -- \\
\midrule
\multicolumn{10}{@{}l}{\textbf{\textit{Ours}}} \\
TLive-Omni & 4B & \textbf{77.7} & 86.6 & 86.6 & \underline{80.5} & 61.3 & 87.4 & 42.3 & 79.3 \\
TLive-Omni & 9B & 76.0 & \textbf{88.6} & \underline{90.3} & \textbf{81.3} & \underline{63.1} & \textbf{90.0} & \textbf{48.0} & \textbf{80.4} \\
\bottomrule
\end{tabular}%
}
\caption{General image benchmark results on HallusionBench, AI2D, OCRBench, CC-OCR, CharXiv(RQ), RefCOCO, ERQA, and EmbSpatialBench. A dash denotes an unreported result or undisclosed parameter count. The Best results among open-source models are marked in \textbf{bold}, while the second-best results are in \underline{underlined}.}
\label{tab:general-image-ocr-results}
\end{table}

Table~\ref{tab:general-video-results} shifts the comparison to video understanding. MVBench~\citep{mvbench} focuses on temporal understanding, MLVU~\citep{mlvu}, LongVideoBench~\citep{longvideobench}, and LVBench~\citep{lvbench} target long-context video reasoning, Video-MME (without subtitles)~\citep{videomme} provides broad-coverage video question answering, while MMVU~\citep{mmvu} and VideoMMMU~\citep{videommmu} measure expert-level knowledge-intensive video understanding across multiple disciplines. TLive-Omni remains competitive, with the 9B model leading open-source models on MLVU, Video-MME, LongVideoBench, and MMVU, and the 4B model leading on VideoMMMU.

\newpage
Table~\ref{tab:timelens-results} reports video temporal grounding performance on TimeLens-Bench~\citep{timelens}. It includes Charades-TL, ActivityNet-TL, and QVHighlights-TL, which evaluate temporal grounding across short daily-life videos, longer activity videos, and mixed-domain videos, respectively. TLive-Omni-4B achieves the highest open-source mIoU on all three benchmarks, while TLive-Omni-9B ranks second on Charades-TL and ActivityNet-TL. These results show that TLive-Omni retains strong general video capability across broad video reasoning, long-context understanding, question answering, and fine-grained temporal grounding.

Table~\ref{tab:general-omni-results} evaluates joint audio-video perception and reasoning with omni-modal benchmarks. AVUT~\citep{avut} evaluates audio-centric video understanding without text shortcuts, DailyOmni~\citep{dailyomni} focuses on audio-visual reasoning with temporal alignment, WorldSense~\citep{worldsense} evaluates real-world omnimodal understanding across visual, audio, and text inputs, and VideoHolmes~\citep{videoholmes} focuses on complex video reasoning. OmniVideoBench~\citep{omnivideobench} evaluates synergistic audio-visual understanding with an emphasis on modality complementarity, while FutureOmni~\citep{futureomni} measures future-event forecasting. Among open-source models, TLive-Omni-9B achieves the best results on AVUT, WorldSense, DailyOmni, and FutureOmni, and ranks second on VideoHolmes and OmniVideoBench. These results indicate that TLive-Omni preserves strong general omni-modal capability across audio-centric video understanding, audio-visual temporal alignment, real-world omnimodal reasoning, and future-oriented video understanding.

\clearpage
\begin{table}[H]
\centering
\tiny
\setlength{\tabcolsep}{2.1pt}
\resizebox{\textwidth}{!}{%
\begin{tabular}{@{}lcccccccc@{}}
\toprule
\multicolumn{1}{c}{\textbf{Model}} & \textbf{Params} & \textbf{MVBench} & \textbf{MLVU} & \textbf{Video-MME} & \textbf{LongVideoBench} & \textbf{LVBench} & \textbf{MMVU} & \textbf{VideoMMMU} \\
\midrule
\multicolumn{9}{@{}l}{\textbf{\textit{Closed-source models}}} \\
Gemini 2.5 Flash & - & -- & 77.8 & 75.6 & -- & 62.2 & 68.2 & 65.2 \\
Gemini 2.5 Pro & - & 65.8 & 81.2 & 80.6 & -- & 69.0 & 72.2 & 79.4 \\
Gemini 3 Pro & - & 74.1 & 83.0 & 87.7 & 76.7 & 76.2 & 77.5 & 87.6 \\
GPT-4o & - & -- & -- & 71.9 & -- & -- & -- & -- \\
GPT-5 (minimal) & - & 64.6 & 78.3 & 77.3 & -- & -- & 68.1 & 61.6 \\
Qwen3.5-Omni Flash & - & 70.8 & 81.9 & 77.0 & -- & -- & 62.7 & -- \\
\midrule
\multicolumn{9}{@{}l}{\textbf{\textit{Open-source VLM models}}} \\
MiMo-VL-SFT & 7B & -- & -- & 66.9 & -- & -- & -- & 53.1 \\
SAIL-VL2 & 8B & -- & -- & 62.7 & 58.3 & -- & -- & -- \\
LLaVA-OneVision-2 & 8B & 66.2 & 76.6 & \underline{71.9} & 66.9 & 55.5 & 56.2 & -- \\
LLaVA-Video & 7B & 58.6 & 70.8 & 63.3 & 58.2 & 44.2 & 47.1 & 36.1 \\
InternVL3.5 & 4B & 71.2 & 70.4 & 65.4 & 60.8 & 43.2 & 47.6 & 57.6 \\
InternVL3.5 & 8B & 72.1 & 70.2 & 66.0 & 62.1 & 46.7 & 60.2 & -- \\
MiniCPM-V 4.5 & 8B & -- & 75.1 & 67.9 & 63.9 & 50.4 & 58.9 & 57.1 \\
LongVU & 7B & 66.9 & 65.4 & 60.6 & -- & -- & -- & -- \\
LongVILA & 7B & 67.1 & -- & 60.1 & 57.1 & -- & -- & -- \\
Mage-VL & 4B & 65.1 & 68.7 & 64.0 & 61.3 & 41.8 & -- & -- \\
Molmo2 & 4B & 75.1 & 63.0 & 69.6 & \underline{68.0} & 53.9 & 51.2 & 50.7 \\
Molmo2 & 8B & \textbf{75.9} & 60.2 & 69.9 & 67.5 & 52.8 & -- & -- \\
NVILA & 8B & 68.1 & 70.1 & 64.2 & 57.7 & -- & -- & -- \\
Kangaroo & 8B & 61.1 & 61.0 & 56.0 & 54.8 & 39.4 & -- & -- \\
Video-XL2 & 8B & -- & 74.8 & 66.6 & 61.0 & 48.4 & 50.0 & 39.9 \\
VideoChat3 & 4B & -- & -- & 70.1 & -- & 56.7 & 56.4 & 57.4 \\
VideoLLaMA 3 & 7B & 69.7 & 73.0 & 66.2 & 59.8 & 45.3 & 44.1 & 34.6 \\
Qwen3-VL & 4B & 68.9 & 75.3 & 69.3 & -- & 56.2 & 50.5 & 56.2 \\
Qwen3-VL & 8B & 68.7 & 78.1 & 71.4 & -- & 58.0 & 58.7 & 65.3 \\
Qwen3.5 & 4B & 66.6 & 75.1 & 71.6 & 65.1 & 55.3 & 57.8 & 69.8 \\
Qwen3.5 & 9B & \underline{75.7} & \underline{79.7} & 66.9 & 67.9 & \textbf{60.9} & \underline{63.7} & 70.3 \\
\midrule
\multicolumn{9}{@{}l}{\textbf{\textit{Open-source Omni models}}} \\
InteractiveOmni & 4B & -- & 68.0 & 63.3 & 57.0 & -- & -- & -- \\
InteractiveOmni & 8B & -- & 71.6 & 66.0 & 59.1 & -- & -- & -- \\
VITA-1.5 & 7B & 55.4 & -- & 56.1 & -- & -- & -- & -- \\
Valley3 & 8B & -- & 55.6 & -- & -- & -- & -- & 61.2 \\
OmniVinci & 9B & 70.6 & -- & 68.2 & 61.3 & -- & -- & -- \\
Nemotron 3 Nano Omni & 30B-A3B & -- & -- & 70.8 & -- & -- & -- & -- \\
Ming-Lite-Omni v1.5 & 20B-A3B & 69.4 & -- & 67.1 & 59.5 & -- & -- & -- \\
MiniCPM-o 2.6 & 8B & -- & -- & 63.9 & -- & -- & -- & -- \\
MiniCPM-o 4.5 & 9B & -- & 76.5 & 70.4 & 66.0 & -- & -- & -- \\
Qwen2.5-Omni & 7B & 70.3 & -- & 64.3 & -- & -- & -- & -- \\
Qwen3-Omni & 30B-A3B & -- & 75.2 & 70.5 & -- & -- & -- & -- \\
\midrule
\multicolumn{9}{@{}l}{\textbf{\textit{Ours}}} \\
TLive-Omni & 4B & 69.0 & 76.1 & 71.3 & 66.1 & 57.1 & 59.9 & \textbf{73.9} \\
TLive-Omni & 9B & 72.5 & \textbf{80.9} & \textbf{75.6} & \textbf{69.9} & \underline{60.8} & \textbf{67.1} & \underline{72.8} \\
\bottomrule
\end{tabular}%
}
\caption{General video benchmark results on MVBench, MLVU, Video-MME, LongVideoBench, LVBench, MMVU, and VideoMMMU. A dash denotes an unreported result or undisclosed parameter count. The Best results among open-source models are marked in \textbf{bold}, while the second-best results are in \underline{underlined}.}
\label{tab:general-video-results}
\end{table}

Taken together, the live-commerce and general benchmark evaluations show that TLive-Omni provides consistent multimodal understanding across audio, image, and video inputs. On live-commerce benchmarks, the models achieve strong results in ASR and speaker-attributed ASR, product visual grounding and text understanding, temporal grounding, dense video caption, video question answering, and shot understanding. These results cover perception and reasoning tasks that require temporal, visual, and audio evidence. On general benchmarks, TLive-Omni remains competitive in image-centric reasoning and question answering, hallucination and OCR evaluation, visual and spatial grounding, long-context video understanding, temporal grounding, and omni-modal perception and reasoning. Across the two model sizes, the 9B variant achieves the best open-source results on many of the reported metrics, while the 4B variant also obtains the best or second-best open-source results across multiple benchmarks. This overall pattern indicates that the strong performance on live-commerce tasks is accompanied by broad performance across general multimodal benchmarks rather than remaining limited to the target domain.

\begin{table}[!htbp]
\centering
\small
\setlength{\tabcolsep}{6pt}
\begin{tabular*}{\textwidth}{@{\extracolsep{\fill}}lcccc@{}}
\toprule
\multicolumn{1}{c}{\textbf{Model}} & \textbf{Params} & \textbf{Charades-TL} & \textbf{ActivityNet-TL} & \textbf{QVHighlights-TL} \\
\midrule
\multicolumn{5}{@{}l}{\textbf{\textit{Closed-source models}}} \\
Gemini 2.5 Flash & - & 48.6 & 52.5 & 64.3 \\
Gemini 2.5 Pro & - & 52.8 & 58.1 & 70.4 \\
GPT-4o & - & 41.8 & 40.4 & 52.1 \\
GPT-5 (minimal) & - & 40.5 & 42.9 & 56.8 \\
\midrule
\multicolumn{5}{@{}l}{\textbf{\textit{Open-source VLM models}}} \\
MiMo-VL-SFT & 7B & 39.6 & 35.5 & 41.5 \\
LLaVA-OneVision-2 & 8B & 53.5 & 53.8 & 66.4 \\
LLaVA-Video & 7B & 15.2 & 14.6 & 10.4 \\
InternVL3.5 & 4B & 16.0 & 14.9 & 17.7 \\
InternVL3.5 & 8B & 27.8 & 31.3 & 31.3 \\
MiniCPM-V 4.5 & 8B & 31.9 & 32.3 & 46.1 \\
Mage-VL & 4B & 50.7 & 45.4 & 57.4 \\
Molmo2 & 4B & 33.3 & 39.8 & 58.7 \\
Video-XL-2 & 8B & 38.9 & 30.0 & 46.2 \\
VideoChat3 & 4B & 56.1 & 54.6 & \underline{67.0} \\
VideoLLaMA 3 & 7B & 39.8 & 29.8 & 36.9 \\
Qwen3-VL & 4B & 46.4 & 48.2 & 58.7 \\
Qwen3-VL & 8B & 48.3 & 46.8 & 59.4 \\
Qwen3.5 & 4B & 48.7 & 51.6 & 55.0 \\
Qwen3.5 & 9B & 52.0 & 54.0 & 57.2 \\
\midrule
\multicolumn{5}{@{}l}{\textbf{\textit{Ours}}} \\
TLive-Omni & 4B & \textbf{57.0} & \textbf{58.2} & \textbf{69.2} \\
TLive-Omni & 9B & \underline{56.3} & \underline{55.4} & 64.1 \\
\bottomrule
\end{tabular*}
\caption{Temporal grounding results on TimeLens-Bench, reported as mIoU on Charades-TL, ActivityNet-TL, and QVHighlights-TL. A dash denotes an undisclosed parameter count. The Best results among open-source models are marked in \textbf{bold}, while the second-best results are in \underline{underlined}.}
\label{tab:timelens-results}
\end{table}
\FloatBarrier

\begin{table}[!htbp]
\centering
\small
\setlength{\tabcolsep}{3pt}
\resizebox{\textwidth}{!}{%
\begin{tabular}{@{}lcccccccc@{}}
\toprule
\multicolumn{1}{c}{\textbf{Model}} & \textbf{Params} & \textbf{AVUT} & \textbf{WorldSense} & \textbf{VideoHolmes} & \textbf{DailyOmni} & \textbf{OmniVideoBench} & \textbf{FutureOmni} \\
\midrule
\multicolumn{8}{@{}l}{\textbf{\textit{Closed-source models}}} \\
Gemini 2.5 Flash & - & 65.4 & 50.9 & -- & -- & -- & 55.6 \\
Gemini 3.1 Pro & - & 85.6 & 65.5 & -- & 82.7 & -- & -- \\
Qwen3.5-Omni Flash & - & 81.4 & 57.9 & -- & 81.8 & -- & -- \\
\midrule
\multicolumn{8}{@{}l}{\textbf{\textit{Open-source Omni models}}} \\
video-SALMONN 2+ & 3B & 66.2 & 48.3 & 42.2 & 67.7 & -- & -- \\
video-SALMONN 2+ & 7B & 69.5 & 50.9 & 46.9 & 71.8 & -- & -- \\
OmniVinci & 9B & -- & 48.2 & -- & 66.5 & 36.7 & 52.8 \\
Nemotron 3 Nano Omni & 30B-A3B & -- & 55.2 & -- & 74.5 & -- & -- \\
MiniCPM-o 4.5 & 9B & \underline{78.6} & \underline{55.7} & \textbf{64.3} & \underline{80.2} & 41.1 & 56.1 \\
Qwen2.5-Omni & 7B & -- & 45.4 & -- & 62.4 & 36.5 & 48.9 \\
Qwen3-Omni & 30B-A3B & 74.2 & 54.0 & 50.4 & 71.9 & \textbf{43.8} & 53.4 \\
\midrule
\multicolumn{8}{@{}l}{\textbf{\textit{Ours}}} \\
TLive-Omni & 4B & \underline{78.6} & 54.0 & 57.5 & 78.6 & 41.6 & \underline{57.2} \\
TLive-Omni & 9B & \textbf{80.0} & \textbf{56.0} & \underline{59.3} & \textbf{80.5} & \underline{43.2} & \textbf{58.5} \\
\bottomrule
\end{tabular}%
}
\caption{General Omni benchmark results on AVUT, WorldSense, VideoHolmes, DailyOmni, OmniVideoBench, and FutureOmni. A dash denotes an unreported result or undisclosed parameter count. The Best results among open-source models are marked in \textbf{bold}, while the second-best results are in \underline{underlined}.}
\label{tab:general-omni-results}
\end{table}

\subsection{Qualitative Analysis}
\label{sec:qualitative-analysis}

\begin{figure}[!t]
\centering
\captionsetup{skip=2pt}
\includegraphics[width=0.98\textwidth,trim=11bp 7bp 8bp 4bp,clip]{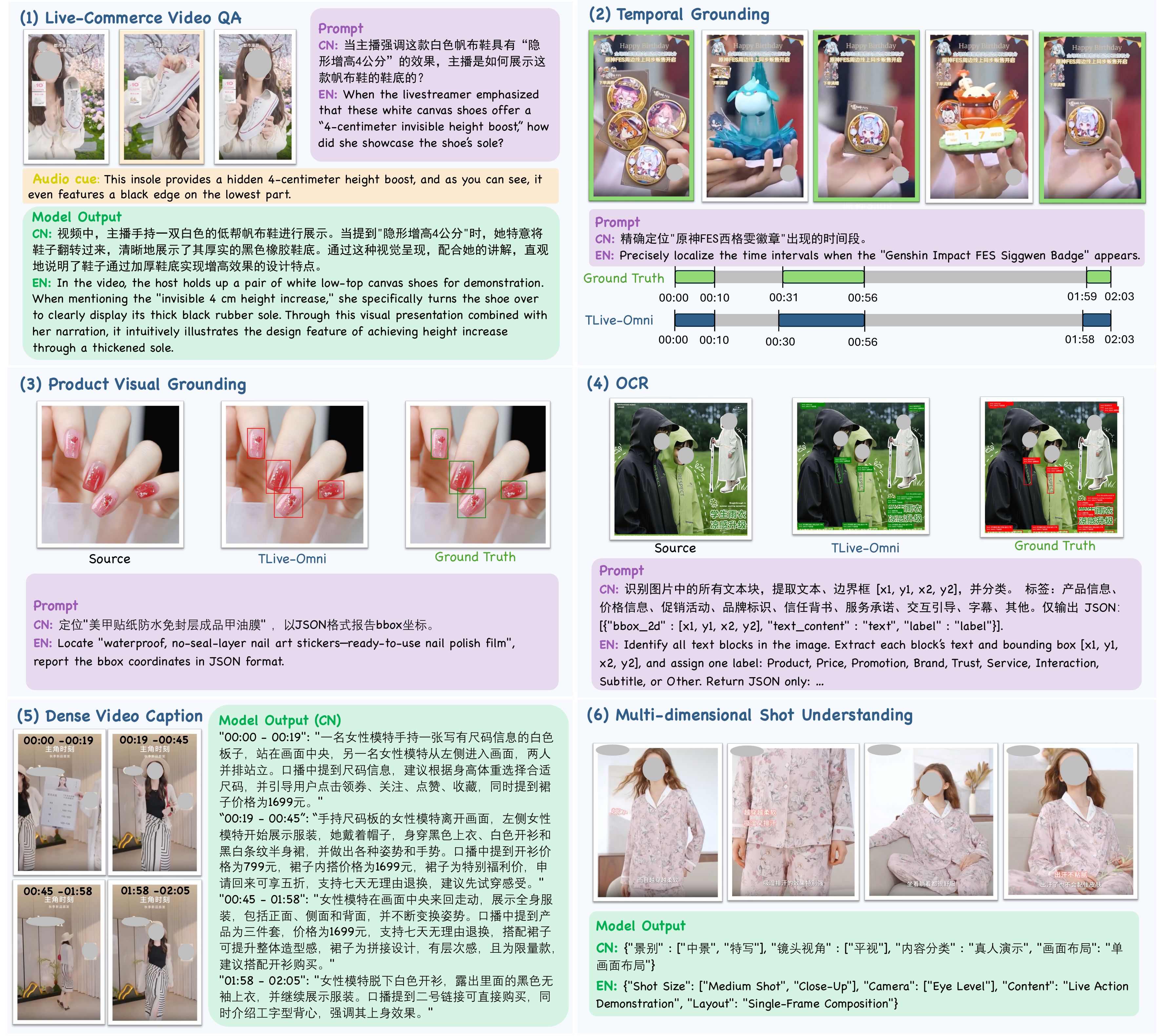}
\caption{Qualitative live-commerce examples of TLive-Omni.}
\label{fig:qualitative-analysis}
\vspace{2pt}
\includegraphics[width=0.98\textwidth,trim=13bp 3bp 6bp 1bp,clip]{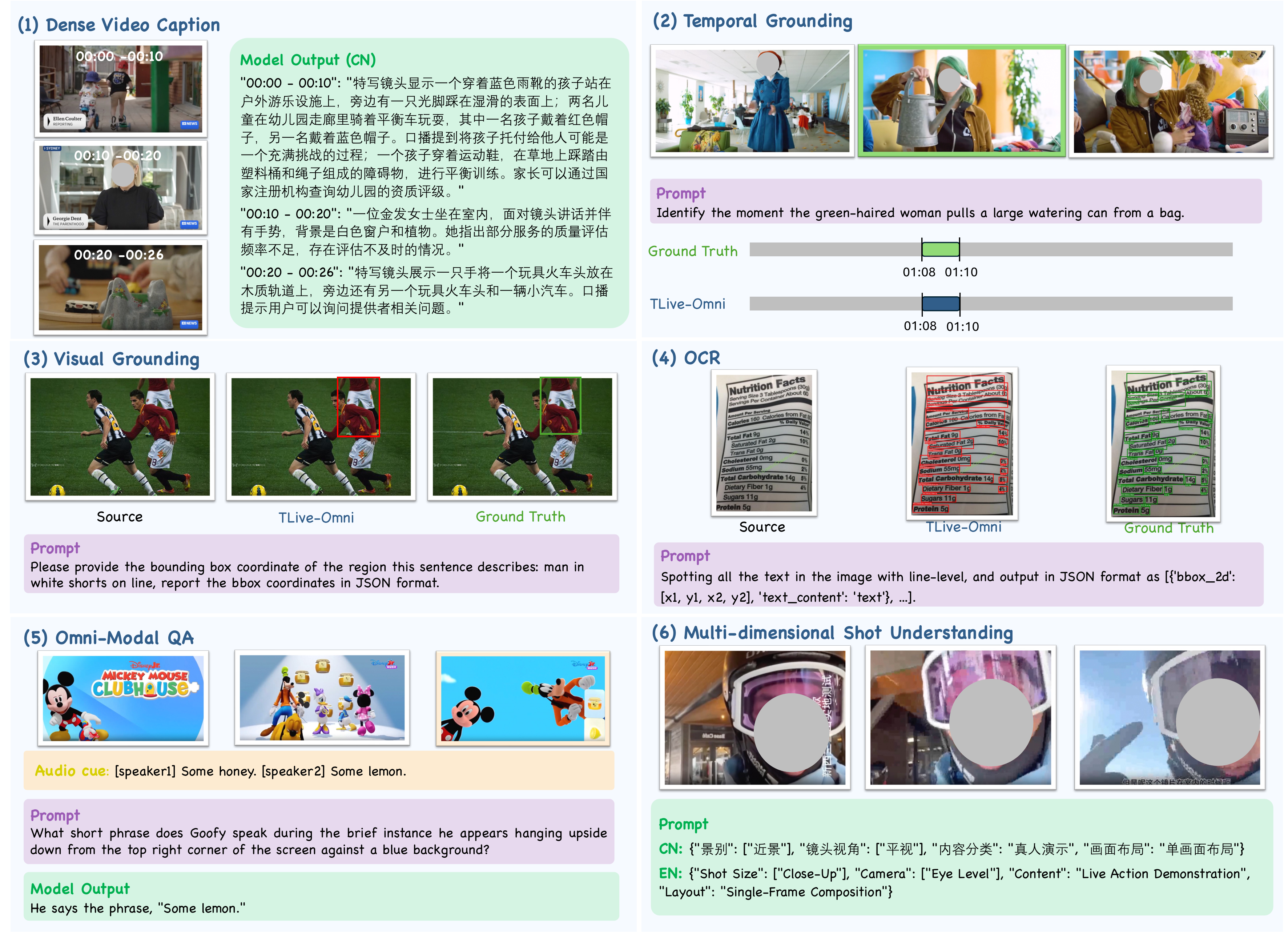}
\caption{Qualitative general-capability examples of TLive-Omni.}
\label{fig:qualitative-general-analysis}
\end{figure}

We present qualitative examples of TLive-Omni in both domain-specific and general settings. Figure~\ref{fig:qualitative-analysis} shows live-commerce examples, while Figure~\ref{fig:qualitative-general-analysis} presents general-capability examples. Together, these cases illustrate TLive-Omni's multimodal understanding across specialized and open-domain scenarios.

\FloatBarrier

\section{Conclusion and Limitations}
We present TLive-Omni, a scenario-oriented omni-modal understanding model for e-commerce live streaming. It maps image, video, audio, and text inputs into a unified representation space, uses Per-vGrid to organize timestamped audio--video evidence, and combines a three-stage supervised training recipe with Faithful-RFT over an atomic live-commerce capability taxonomy. The live-commerce evaluation shows strong performance across speech transcription and speaker-attributed ASR, audio description and question answering, product visual grounding, text localization/recognition/classification, temporal grounding, dense video captioning, video question answering, and shot understanding. The general benchmark results further show that TLive-Omni retains broad multimodal capabilities beyond the target domain, with competitive results across image, video and omni-modal evaluations, and improvements over the corresponding Qwen3.5 4B and 9B backbones on multiple benchmarks. Overall, these results suggest that aligning model design, data construction, training objectives, and evaluation protocols with the target deployment scenario is important for building a strong omni-modal model for e-commerce live streaming.

Despite these results, TLive-Omni remains focused on understanding rather than generation or full-duplex real-time interaction. Future work will extend evaluation coverage on broader public benchmarks and further improve robustness for longer, noisier, and more diverse live-stream scenarios. Another direction is to strengthen the calibration of temporal evidence under incomplete or ambiguous multimodal inputs, which are common in practical live-stream settings.

\section*{Contributors}
\textbf{Project Lead:} Yibo Hu.\\
\textbf{Contributors:} Yu Qian, Mao Gu, Yingfan Tao, Yuhao Chen, Yongdong Luo, Zhuoqun Liu, Yibo Hu, Meiguang Jin, Junfeng Ma. \\
\textbf{Contact:} huyibo871079699@gmail.com

\clearpage
\appendix
\section*{Appendix}
\section{Related Work}
\label{app:related-work}

\subsection{Omni-Modal Large Models}
\label{app:related-omni}

Recent omni-modal models have shown that image, video, audio and text can be integrated into a shared language-model interface. Gemini~\citep{gemini} and GPT-4~\citep{gpt4} demonstrate the effectiveness of large-scale multimodal systems, while open omni-modal models such as Qwen2.5-Omni~\citep{qwen25omni}, Qwen3-Omni~\citep{qwen3omni}, Qwen3.5-Omni~\citep{qwen35omni}, Baichuan-Omni-1.5~\citep{baichuanomni15}, OmniVinci~\citep{omnivinci}, MiniCPM-o 4.5~\citep{minicpmo45}, and Nemotron 3 Nano Omni~\citep{nemotron3nano} make this direction increasingly accessible. These models provide the architectural foundation for unified multimodal interaction, but they are primarily organized around open-domain capabilities rather than the long-form, product-centric, and temporally grounded demands of e-commerce live streaming.

\subsection{E-commerce and Live-stream Multimodal Understanding}
\label{app:related-ecommerce}

E-commerce multimodal research has developed along several complementary paths. The MOON series~\citep{moon,moon2,moon3} focuses on product representation learning from multimodal product content, while E-VAds~\citep{evads} introduces a benchmark for evaluating commercial-intent reasoning in e-commerce short videos. Valley3~\citep{valley3} extends e-commerce modeling toward an omni foundation model, and LiViBench~\citep{livibench} highlights the evaluation challenges of interactive live-stream videos. These works motivate our setting, but TLive-Omni targets a different emphasis: a live-commerce understanding model whose architecture, data construction, and capability taxonomy are organized around joint audio, video, image, text and product-grounded evidence.

\subsection{Reinforcement Learning for Large Language Models}
\label{app:related-rl}

GRPO~\citep{deepseekmath} and related verifiable-reward methods have made reinforcement learning (RL) post-training a practical strategy for improving reasoning-oriented models. For vision-language models, recent RL studies directly optimize perceptual correctness in visual understanding, hallucination mitigation, and temporal grounding~\citep{perceptionr1,papo,pdcr,disentangling,omnidpo,timer1}. GRPO has also been applied to speech recognition~\citep{grpoasr}. These directions motivate a broader view of multimodal post-training: rewards should evaluate whether a response is supported by modality-specific evidence, rather than reasoning alone. Faithful-RFT follows this perception-centered view: it uses task-conditioned reward routing over image, video, and audio streams to score final-answer quality, without rewarding long chain-of-thought or treating reasoning traces as the objective.

\section{Evaluation Metrics}
\label{app:evaluation-metrics}

We use standard accuracy, AP, F1, and CER definitions unless otherwise noted. For ASR, both the ground-truth transcript and the model-generated transcript are first normalized by removing speaker markers, bracketed tags, punctuation, spaces, and modal particles, and by converting Chinese text to simplified Chinese. CER is then computed as $(S+D+I)/N$, where $S$, $D$, and $I$ are the numbers of character substitutions, deletions, and insertions, and $N$ is the number of characters in the ground-truth transcript.

\paragraph{cpWER.}
Speaker-attributed ASR is evaluated using concatenated minimum-permutation WER, following the meeting-transcription convention~\citep{meeteval}. WER is the word-level error rate, computed from word substitutions, deletions, and insertions relative to the ground-truth transcript. All transcript segments assigned to the same speaker are first concatenated separately for the ground-truth and model predictions. Predicted speakers are then matched to ground-truth speakers using the assignment that minimizes the total word error. If the two sides contain different numbers of speakers, empty streams are added to the smaller side before matching. cpWER is the WER after this optimal speaker pairing.

\paragraph{OCR and grounding.}
Product visual grounding uses AP under an IoU threshold of 0.5 with one-to-one matching between predicted and ground-truth boxes. OCR localization uses an IoU threshold of 0.5 to match predicted and ground-truth text boxes one to one, and computes F1 from the matched boxes. OCR recognition is evaluated on IoU-matched text boxes: predicted and ground-truth boxes are matched one to one using $\mathrm{IoU} > 0.5$, and the lower-is-better normalized edit distance $d_{\mathrm{edit}}/\max(|p|,|g|)$ is computed for each matched text pair after normalizing whitespace, where $d_{\mathrm{edit}}$ is the edit distance between the predicted text $p$ and the ground-truth text $g$.

\paragraph{Temporal and video metrics.}
For temporal grounding, mIoU averages interval IoU over samples, with $\mathrm{IoU}_i=|P_i\cap G_i|/|P_i\cup G_i|$ for predicted interval $P_i$ and ground-truth interval $G_i$, where $|\cdot|$ denotes interval length. For multi-interval examples, we first match individual predicted intervals to ground-truth intervals, compute IoU for each matched pair, and then average these IoUs. 
Audio description and dense video caption are evaluated through a caption-based question-answering protocol. For each audio or video sample, we combine model-assisted question generation and verification with human review to construct multiple-choice questions grounded in modality-specific reference annotations. Audio questions focus on product attributes, prices and promotions, and purchase or interaction instructions, while video questions cover visual and spoken content as well as bidirectional temporal grounding. Together, they probe entities and attributes, actions, scenes, events, and temporal relations. Each question contains one annotation-supported answer and plausible distractors derived from confusable or unsupported content, yielding three to five options including ``cannot determine.'' At test time, the evaluated model generates a description or caption from the original audio or video. The generated text and preconstructed questions are then passed to a separate evaluator LLM, which answers solely from the generated text without access to the original input. The evaluator's answers are compared with the ground-truth answers, and accuracy is the fraction that are correct. Hallucination rate is the fraction of incorrect answers among valid questions for which the evaluator LLM selects a concrete answer rather than ``cannot determine.'' 
For shot understanding, the model predicts four structured tags for each clip: layout, shot size, camera angle, and content category. We report accuracy separately for each dimension. A single-choice tag must exactly match the ground-truth tag, while for multi-choice tags, a predicted subset of the ground-truth set receives partial credit of 0.5, and wrong or extra tags receive 0.

\section{Additional In-Context ASR Results}
\label{app:in-context-asr}

In-Context ASR evaluates whether a model can use domain-specific keyword prompts to improve transcription of product names, brand names, and other domain terms. Unlike standard ASR, which transcribes audio without textual hints, this setting provides a candidate keyword list before transcription. The keyword-list size ranges from 0 to 1000. The zero-keyword setting serves as the no-context baseline, while larger lists test whether additional context improves keyword recognition or introduces interference. We report keyword recall and CER: the former measures whether target keywords are recovered in the transcription, while the latter measures the overall character error rate, capturing whether keyword prompting affects the full transcript beyond the target terms. We compare TLive-Omni with Qwen3-ASR-Flash~\citep{qwen3asr,qwen3asrflash} and Qwen3-Omni~\citep{qwen3omni}. As shown in Table~\ref{tab:in-context-asr}, keyword prompting substantially improves recall for both TLive-Omni variants while reducing their CER. TLive-Omni-9B achieves the lowest CER at every nonzero keyword-list size and the highest recall for lists containing 200--1000 keywords.

\begin{table}[t]
\centering
\footnotesize
\setlength{\tabcolsep}{2.5pt}
\begin{tabular*}{\textwidth}{@{\extracolsep{\fill}}lcccccccc@{}}
\toprule
\textbf{Keyword} & \multicolumn{2}{c}{\textbf{Qwen3-ASR-Flash}} & \multicolumn{2}{c}{\textbf{Qwen3-Omni}} & \multicolumn{2}{c}{\textbf{TLive-Omni-4B}} & \multicolumn{2}{c}{\textbf{TLive-Omni-9B}} \\
\cmidrule(lr){2-3}\cmidrule(lr){4-5}\cmidrule(lr){6-7}\cmidrule(lr){8-9}
\textbf{Count} & \textbf{Recall $\uparrow$} & \textbf{CER $\downarrow$} & \textbf{Recall $\uparrow$} & \textbf{CER $\downarrow$} & \textbf{Recall $\uparrow$} & \textbf{CER $\downarrow$} & \textbf{Recall $\uparrow$} & \textbf{CER $\downarrow$} \\
\midrule
\textbf{0} & \underline{51.88} & \textbf{6.33} & \textbf{53.44} & 6.59 & 49.12 & 6.69 & 48.62 & \underline{6.54} \\
\textbf{50} & 75.69 & \underline{5.40} & \textbf{85.44} & 7.64 & 79.06 & 5.41 & \underline{81.00} & \textbf{5.18} \\
\textbf{100} & 76.94 & 6.59 & \textbf{81.56} & 7.34 & 77.75 & \underline{5.45} & \underline{80.31} & \textbf{5.19} \\
\textbf{200} & 68.00 & 5.73 & \underline{78.69} & 8.13 & 76.56 & \underline{5.56} & \textbf{79.50} & \textbf{5.30} \\
\textbf{300} & 67.12 & 5.89 & \underline{76.94} & 7.65 & 76.06 & \underline{5.61} & \textbf{78.31} & \textbf{5.36} \\
\textbf{500} & 60.56 & 6.07 & 74.38 & 7.49 & \underline{74.94} & \underline{5.71} & \textbf{77.31} & \textbf{5.42} \\
\textbf{1000} & 56.81 & 10.68 & 70.56 & 7.54 & \underline{72.44} & \underline{5.76} & \textbf{74.31} & \textbf{5.53} \\
\bottomrule
\end{tabular*}
\caption{In-Context ASR results under different keyword-list sizes. Recall denotes keyword recall, and CER denotes character error rate. For each keyword-list size, best values are shown in \textbf{bold} and second-best values are \underline{underlined}.}
\label{tab:in-context-asr}
\end{table}

\section{Training and Faithful-RFT Details}
\label{app:faithful-rft-details}

\subsection{Three-Stage SFT Hyperparameters}
\label{app:sft-hparams}

The three SFT stages introduced in Section~\ref{sec:training-recipe} (audio-language alignment, audio strengthening, and full multimodal SFT) all use the AdamW optimizer with $\beta_1=0.9$, $\beta_2=0.95$, a cosine learning-rate schedule, a weight decay of 0.1, gradient clipping at a maximum norm of 1, ZeRO-3 sharding via DeepSpeed~\citep{zero,deepspeed}, and gradient checkpointing. Each stage is trained for a single epoch over its stage-specific data mixture.
The global batch size is 1,024 for Stage 1, 2,048 for Stage 2, and 1,024 for Stage 3. The learning rate is $1\times10^{-4}$, $1\times10^{-5}$, and $4\times10^{-6}$ for the three stages, respectively, and the warmup ratio is 0.01, 0.01, and 0.05.

\subsection{Faithful-RFT Hyperparameters}

For Faithful-RFT, the number of candidate responses per prompt is set to $G=8$. Group-relative rewards are normalized with a smoothing constant of $\epsilon_s=10^{-4}$. The clipped objective uses $\epsilon_l=0.2$ and $\epsilon_h=0.28$ for the lower and upper policy-ratio bounds, respectively, together with a KL coefficient of $\beta=0.1$.

\section{Prompts for General Benchmark Evaluation}
\label{app:general-benchmark-prompts}

This section reports the exact prompts used to evaluate TLive-Omni on the general-purpose benchmarks introduced in Section~\ref{sec:general-benchmark-evaluation}. Prompts are grouped by input modality.

\subsection{Image Benchmarks}
\label{app:general-prompts-image}

{\footnotesize
\renewcommand{\arraystretch}{1.15}
\setlength{\tabcolsep}{5pt}
\begin{longtable}{>{\raggedright\arraybackslash}p{0.20\textwidth}>{\raggedright\arraybackslash\fontfamily{pcr}\selectfont\footnotesize}p{\textwidth-0.20\textwidth-4\tabcolsep}}
\toprule
\rowcolor{white}
\normalfont Benchmark(s) & \normalfont Prompt Template \\
\midrule
\endfirsthead
\toprule
\rowcolor{white}
\normalfont Benchmark(s) & \normalfont Prompt Template \\
\midrule
\endhead
\midrule
\endfoot
\bottomrule
\endlastfoot
\rowcolor{gray!10}
MMMU, MMBench, MMStar, AI2D
 & \{question\} \newline Options: \newline
 A. \{option\_a\} / B. \{option\_b\} / C. \{option\_c\} / D. \{option\_d\} \newline
 Think step by step before answering. The last line of your response should be of the following format: `Answer: \$LETTER' (without quotes) where LETTER is one of the options. \\
VLMsAreBlind, RealWorldQA, ERQA, EmbSpatialBench
 & Hint: \{hint\} \newline Question: \{question\} \newline Options: \newline
 A. \{option\_a\} / B. \{option\_b\} / ... \newline
 Please select the correct answer from the options above. \\
\rowcolor{gray!10}
MathVista
 & \{hint\} \newline
 \{question\} \newline
 Think step by step before answering. The last line of your response should be of the following format: `Answer: \$ANSWER' (without quotes) where \$ANSWER is your final answer. \\
CharXiv, SimpleVQA
 & \{question\} \newline
 Think step by step before answering. The last line of your response should be of the following format: `Answer: \$ANSWER' (without quotes) where \$ANSWER is your final answer. \\
\rowcolor{gray!10}
DynaMath
 & \#\# Question \newline
 \{question\} \newline
 \newline
 \#\# Answer Instruction Please provide an answer to the question outlined above. Your response should adhere to the following JSON format, which includes two keys: `solution' and `short answer'. The `solution' key can contain detailed steps needed to solve the question, and the `short answer' key should provide a concise response. Provide the corresponding choice option in the `short answer' key, such as `A', `B', `C', or `D'. \newline
 \newline
 Example of expected JSON response format: \newline
 \{``solution'': ``[Detailed step-by-step explanation]'', ``short answer'': ``[Concise Answer]''\} \\
HallusionBench
 & \{question\} \newline
 Think step by step before answering. The last line of your response should be of the following format: `Answer: Yes/No' (without quotes). \\
\rowcolor{gray!10}
OCRBench
 & What is written in the image? \\
CC-OCR
 & Please output only the text content from the image without any additional descriptions or formatting. \\
\rowcolor{gray!10}
RefCOCO
 & Please provide the bounding box coordinate of the region this sentence describes: <ref>\{sentence\}</ref> \\
\end{longtable}
\renewcommand{\arraystretch}{1}
}

\subsection{Video Benchmarks}
\label{app:general-prompts-video}

\begin{center}
\footnotesize
\renewcommand{\arraystretch}{1.15}
\setlength{\tabcolsep}{5pt}
\begin{tabular}{>{\raggedright\arraybackslash}p{0.20\textwidth}>{\raggedright\arraybackslash\fontfamily{pcr}\selectfont\footnotesize}p{\textwidth-0.20\textwidth-4\tabcolsep}}
\toprule
\rowcolor{white}
\normalfont Benchmark(s) & \normalfont Prompt Template \\
\midrule
\rowcolor{gray!10}
MVBench, Video-MME, MLVU, LongVideoBench, LVBench, MMVU
 & \{question\} \newline A. \{option\_a\} / B. \{option\_b\} / ... \newline
 Answer with the option's letter from the given choices directly. \\
VideoMMMU (multiple-choice)
 & \{question\} \newline A. \{option\_a\} / B. \{option\_b\} / ... \newline
 Think step by step before answering. End your response with ``Answer: X'', where X is the option letter. \\
\rowcolor{gray!10}
VideoMMMU (open-ended)
 & \{question\} \newline
 Think step by step before answering. Add ``Answer: \{Your final answer\}'' at the end of your reply. \\
Charades-TL, ActivityNet-TL, QVHighlights-TL
 & Query: ``\{question\}'' \newline Return only one timestamp range in mm:ss--mm:ss format. \\
\bottomrule
\end{tabular}
\renewcommand{\arraystretch}{1}
\end{center}

\subsection{Omni-Modal Benchmarks}
\label{app:general-prompts-omni}

\begin{center}
\footnotesize
\renewcommand{\arraystretch}{1.15}
\setlength{\tabcolsep}{5pt}
\begin{tabular}{>{\raggedright\arraybackslash}p{0.20\textwidth}>{\raggedright\arraybackslash\fontfamily{pcr}\selectfont\footnotesize}p{\textwidth-0.20\textwidth-4\tabcolsep}}
\toprule
\rowcolor{white}
\normalfont Benchmark(s) & \normalfont Prompt Template \\
\midrule
\rowcolor{gray!10}
VideoHolmes, WorldSense, DailyOmni, OmniVideoBench, AVUT, FutureOmni
 & \{question\} \newline A. \{option\_a\} / B. \{option\_b\} / C. \{option\_c\} / D. \{option\_d\} \newline
 Answer with the option's letter from the given choices directly. \\
\bottomrule
\end{tabular}
\renewcommand{\arraystretch}{1}
\end{center}

\clearpage
\bibliography{aaai2027}

\end{document}